\documentclass[twoside,11pt]{article}
\usepackage{theapa,jair,rawfonts}

\jairheading{82}{2025}{1999-2056}{08/2024}{03/2025}
\ShortHeadings{TabID: Automatic Identification and Tabulation of Subproblems}
{Akgün, Gent, Jefferson, Kiziltan, Miguel, Nightingale, Salamon \& Ulrich-Oltean}
\firstpageno{1999}

\usepackage{listings}
\usepackage[utf8]{inputenc}
\usepackage[T1]{fontenc}
\usepackage{amssymb}
\usepackage{amsmath}
\usepackage{booktabs}
\usepackage{graphicx}
\usepackage{xspace}
\usepackage{xcolor}
\definecolor{ourcolsGray}{gray}{0.6}
\definecolor{ourcolsLightGray}{gray}{0.9}
\definecolor{ourcolsLineGray}{gray}{0.3}
\definecolor{ourcolsHighlight}{RGB}{192,0,0}

\usepackage[textsize=footnotesize,colorinlistoftodos]{todonotes}

\usepackage{tikz}
\usetikzlibrary{shapes.geometric,arrows,bending,positioning,fit,patterns,shapes,quotes,decorations.pathreplacing,calc}

\usepackage{courier}

\usepackage[implicit=false,bookmarks=false,breaklinks=true,colorlinks,linkcolor=blue,citecolor=blue,urlcolor=blue,bookmarksnumbered]{hyperref}

\usepackage[nameinlink]{cleveref}

\newcommand{\comment}[1]{}

\newcommand{\savilerow}{\textsc{Savile Row}\xspace}
\newcommand{\eprime}{Essence Prime\xspace}

\newcommand{\tick}{\checkmark}

\lstdefinelanguage{eprime}{
morekeywords={language, Essence, given, letting, find, such, that,
            bool, int, matrix, enum, variant, record, set, mset,
            sequence, function, relation, partition, 
            domain, total, surjective, be, allDiff,
            forAll, exists, sum, injective, in, preImage, range,
            new, type, intersect, union, from,
            minimising, maximising, of, indexed, by, and, or, toInt, numParts, partSize, together,
            defined, maxSize, maxNumParts, size, regular, language},
sensitive=true,
morecomment=[l]{\$}
}

\newcommand{\code}[1]{\lstinline~#1~}
\newcommand{\zap}[1]{}

\begin{document}

\title{TabID: Automatic Identification and Tabulation \\ of Subproblems in Constraint Models}

\author{\name Özgür Akgün \email ozgur.akgun@st-andrews.ac.uk \\
\name Ian P.\ Gent \email Ian.Gent@st-andrews.ac.uk \\
\addr School of Computer Science, University of St Andrews\\
St Andrews, Fife KY16 9SX, UK 
\AND
\name Christopher Jefferson \email cjefferson001@dundee.ac.uk \\
\addr School of Science and Engineering, University of Dundee\\
Dundee, DD1 4HN, UK 
\AND
\name Zeynep Kiziltan \email zeynep.kiziltan@unibo.it \\
\addr Department of Computer Science and Engineering\\
University of Bologna\\
Mura Anteo Zamboni 7, 40126, Bologna, Italy 
\AND
\name Ian Miguel \email ijm@st-andrews.ac.uk \\
\addr School of Computer Science, University of St Andrews\\
St Andrews, Fife KY16 9SX, UK 
\AND
\name Peter Nightingale \email peter.nightingale@york.ac.uk \\
\addr Department of Computer Science, University of York\\
Heslington, York YO10 5GH, UK
\AND
\name Andr\'as Z.\ Salamon \email Andras.Salamon@st-andrews.ac.uk \\
\addr School of Computer Science, University of St Andrews\\
St Andrews, Fife KY16 9SX, UK 
\AND
\name Felix Ulrich-Oltean \email felix.ulrich-oltean@york.ac.uk \\
\addr Department of Computer Science, University of York\\
Heslington, York YO10 5GH, UK
}

\maketitle

\begin{abstract}

The performance of a constraint model can often be improved by converting a subproblem into a single table constraint (referred to as tabulation). Finding subproblems to tabulate is traditionally a manual and time-intensive process, even for expert modellers. This paper presents TabID, an entirely automated method to identify promising subproblems for tabulation in constraint programming. We introduce a diverse set of heuristics designed to identify promising candidates for tabulation, aiming to improve solver performance.
These heuristics are intended to encapsulate various factors that contribute to useful tabulation.
We also present additional checks to limit the potential drawbacks of suboptimal tabulation. 

We comprehensively evaluate our approach using benchmark problems from existing literature that previously relied on manual identification by constraint programming experts of constraints to tabulate.
We demonstrate that our automated identification and tabulation process achieves comparable, and in some cases improved results. We empirically evaluate the efficacy of our approach on a variety of solvers, including standard CP (Minion and Gecode), clause-learning CP (Chuffed and OR-Tools) and SAT solvers (Kissat).

Our findings highlight the substantial potential of fully automated tabulation, suggesting its integration into automated model reformulation tools.

\end{abstract}

\section{Introduction}\label{sec:introduction}

Constraint programming provides an efficient means of solving complex combinatorial problems across a wide variety of disciplines, such as scheduling, planning, routing, and configuration \shortcite{handbook-cp}. In order to solve a problem using the constraint programming paradigm, it must first be {\em modelled} in a format suitable for input to a constraint solver. This involves determining the set of decision variables that represent the choices that must be made to solve the problem, and formulating a set of constraints over the variables so as to allow only valid combinations of decisions. For example, in a scheduling scenario we might employ a decision variable per task to represent the start time of that task, with constraints to disallow a number of simultaneous tasks that would exceed the resources available. 

Once a model has been chosen, a constraint solver automatically searches for a solution: a complete assignment of values to the decision variables that satisfies all of the constraints. Search is interleaved with inference known as {\em constraint propagation}, where deductions are made based on the constraints and the current set of assignments made via search. These deductions serve to narrow down the choices for the variables as yet unassigned by search, and therefore reduce the search required. Generally, there are many ways in which a given problem may be modelled, and the model chosen has a significant effect on the performance of the constraint solver in searching for solutions. Therefore, automated methods for improving constraint models are valuable \shortcite{presolving-minizinc-2015-mod,sr-journal-17}.

In order to improve the performance of a constraint model, a common step is to reformulate the expression of a subset of the problem constraints, either to strengthen the inferences made during search by the constraint solver by increasing constraint propagation, or to maintain the level of propagation while reducing the cost of propagating the constraints. One such method is {\em tabulation}, the aggregation of a set of constraint expressions into a single table constraint~\cite{Mohr1988:good}. Such a table constraint explicitly lists the allowed tuples of values for the decision variables involved. This allows us to exploit efficient table constraint propagators that enforce generalised arc consistency (GAC)~\cite{DBLP:reference/fai/Bessiere06}, typically a stronger level of inference than is achieved for a logically equivalent collection of separate constraints. Successful examples of this approach where the reformulation has been performed by hand include Black Hole patience~\fullcite{Gent2007:search} and Steel Mill Slab Design \cite{gargani2007:steelmill}.

\shortciteA{Dekker2017:autotabling} presented a method for the automation of tabulation (there called `auto-tabling'). In their approach a predicate (a Boolean function) expressed in the MiniZinc language \shortcite{nethercote2007minizinc} may be annotated.
Such an annotation requests that the predicate be converted into a table constraint.  While called `auto-tabling', note that the choice of predicate to be tabulated is done manually by the modeller. IBM ILOG CPLEX Optimization Studio~\cite{ibm2017} and ECLiPSe~\cite{leprovost1992:domain} have similar facilities to generate table constraints, 
while other approaches target alternatives 
such as Multi-valued Decision Diagrams (MDDs)  and regular constraints~\shortcite{Una2018:compiling,Loffler2020:regularization}. In all of these approaches, the crucial first step of identifying promising parts of a given model for tabulation is left to the human modeller. 

In this work we present an entirely automatic tabulation method, TabID, situated in the constraint modelling tool \savilerow \cite{sr-journal-17}. The core function of \savilerow is to translate the constraint modelling language \eprime~\cite{sr-manual} into the input languages of solvers, including constraint programming (CP) and Boolean satisfiability (SAT) solvers. A set of heuristics is employed to identify in an \eprime model some candidate sets of constraints for tabulation, which are then tabulated automatically. Resource limits are applied to this process, ensuring that only the most useful candidate sets of constraints are tabulated. In order to demonstrate the effectiveness of our approach, we first examine the same four case studies that were used by~\citeA{Dekker2017:autotabling} to showcase the utility of tabulation from explicit model annotations. We show that our automated approach can identify the same opportunities to improve the models by tabulation.  We also study nine additional problem classes that show that TabID is effective on a wider range of problems. 
This paper builds on (and entirely includes) an earlier conference publication~\shortcite{cp-tabulation-2018}.

\subsection{Contributions}
The primary contribution of TabID is the automation of a hitherto difficult manual task: the recognition 
of opportunities to tabulate parts of a constraint model in order to increase constraint propagation and therefore reduce search. In support of this primary contribution, we contribute the following:
\begin{itemize}
    \item A set of heuristics to identify common tabulation opportunities, such as expressions that will propagate weakly.
    \item A caching system to avoid tabulating equivalent subproblems multiple times.
    \item A system of progress checks and work limits for the situation where a heuristic identifies a constraint that is too large to be tabulated.
    \item An empirical study across several solvers of the frequency with which our heuristics identify effective tabulation opportunities.
\end{itemize}

\subsection{Motivating Examples}
\label{sec:MotivatingEx}

We consider two motivating examples from the literature: Black Hole and Knight's Tour.

\begin{figure}
    \centering
    \includegraphics[width=0.485\textwidth]{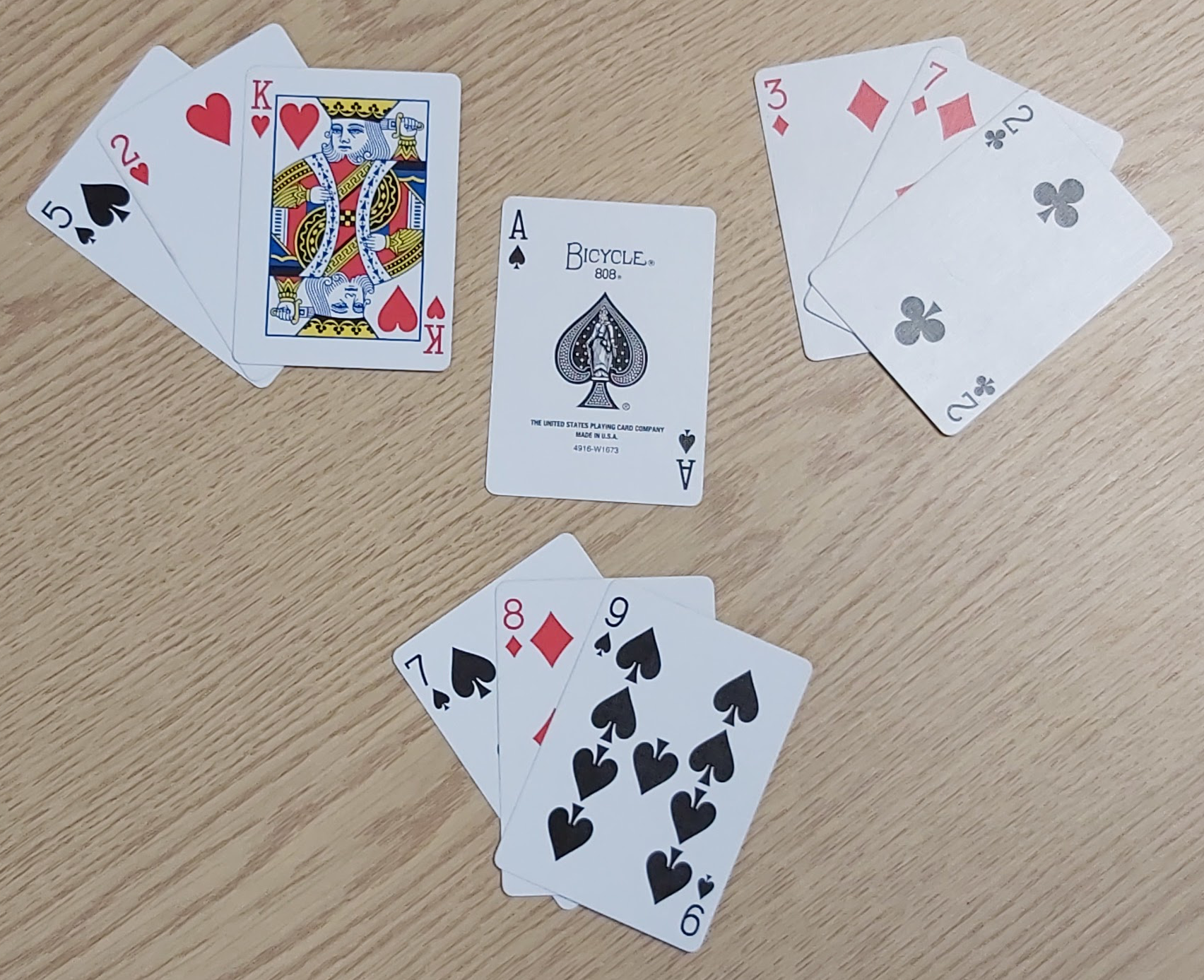}\hfill\includegraphics[width=0.414\textwidth,height=0.396\textwidth]{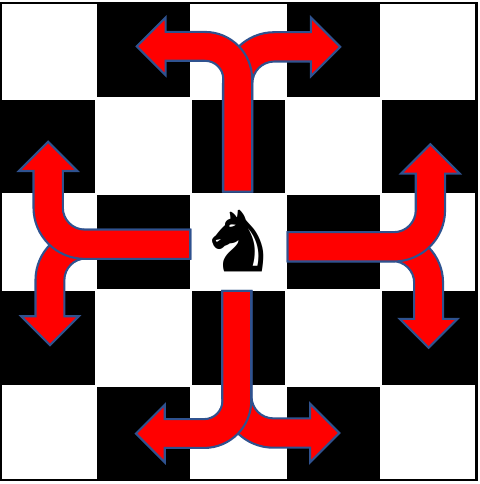}
    \caption{Left: illustration of part of a Black Hole patience setup, showing only 3 (of 17) fans and the `black hole' (Ace of Spades) at the beginning of the game. The King of Hearts and the 2 of Clubs may be played onto the black hole at this point, because they are rank-adjacent to the card on top of the black hole. The 2 of Hearts cannot be played because it is covered. Right: illustration of the moves available for a knight in chess.}
    \label{fig:bh-example}\label{fig:KnightDiagram}
\end{figure}

Black Hole is a single-player card game (variously termed `patience' or `solitaire' depending on the variety of English spoken) where cards are played one by one into the `black hole' from seventeen face-up fans of three cards. All cards can be seen at all times. A card may be played into the black hole if it is adjacent in rank to the previous card. \Cref{fig:bh-example} shows part of a Black Hole game, illustrating the rank adjacency condition. Black Hole was modelled for a variety of solvers by~\fullciteA{Gent2007:search} and a table constraint was used in the CP model.

A constraint model of Black Hole is presented in \Cref{fig:BHModel}. We use the simplest and most declarative model of~\citeA{Dekker2017:autotabling} where two variables \code{a} and \code{b} (with cards coded as integers \(\{0\ldots 51\}\)) represent adjacent cards iff \code{|a-b| \% 13 in \{1,12\}} (the adjacency constraint). The model in the figure uses the available arithmetic and logical operators in the language to capture the rank adjacent condition in the game. The cards in each fan must be played in order. This constraint is captured by ordering variables in the \code{cardSeq} matrix which represents the time step that each card is played. 

The adjacency constraint is a relatively complex expression containing sum, modulo, and absolute value operators. Typically this constraint will propagate weakly because some solvers implement modulo and absolute value poorly, and also (depending on the solver used) the sum operator may implement bound consistency.

\citeA{Dekker2017:autotabling} show that manually tabulating this constraint significantly improves the solving performance. This problem is therefore a good candidate for evaluating our work and indeed we will see that the tabulation opportunities present in this model can be identified and exploited automatically by TabID.

\begin{figure}
\centering
\begin{lstlisting}[language=eprime]
given initialStacks : matrix [int(0..50)] of int(1..51)
find blackHole : matrix [int(0..51)] of int(0..51)
find cardSeq : matrix [int(0..51)] of int(0..51)
such that 
blackHole[0] = 0,  cardSeq[0] = 0,
allDiff(cardSeq),
allDiff(blackHole),
forAll step : int(1..51). 
    (|blackHole[step-1] - blackHole[step]| % 13 in {1,12}),
forAll card : int(0..50). (card % 3 != 2) ->
    (cardSeq[initialStacks[card]] < cardSeq[initialStacks[card+1]]),
forAll step : int(0..51). forAll card : int(0..51). 
    (blackHole[step] = card) <-> (cardSeq[card] = step)
\end{lstlisting}
\caption{A model of Black Hole in the constraint modelling language \eprime \cite{sr-manual}. The parameters to the model (the initial layout of the cards, named \code{initialStacks}) are introduced with the keyword {\tt given}. The decision variables (the single-dimensional matrices {\tt blackHole} and \code{cardSeq}) are introduced with the keyword {\tt find}. Playing cards are numbered \(0\ldots 51\), with spades in the range \(0\ldots 12\), followed by hearts, clubs, and diamonds.}
\label{fig:BHModel}
\end{figure}

Our second motivating example is Knight's Tour, a classic puzzle studied by Euler (1759). More recently, \citeA{Schwenk1991:rectangular} determined the set of board sizes (including rectangular boards) that have a knight's tour. Given a natural number $n$ and a starting square on an $n \times n$ chess board, the task is to find a sequence of moves for the knight visiting each of the remaining squares of the board exactly once. The moves of a knight are illustrated in Figure \ref{fig:KnightDiagram}. We use the Hamiltonian path version of Knight's Tour: the last square visited is not required to be a knight's move from the starting square.

\comment{
\begin{figure}
    \centering
    \includegraphics[width=0.3\textwidth]{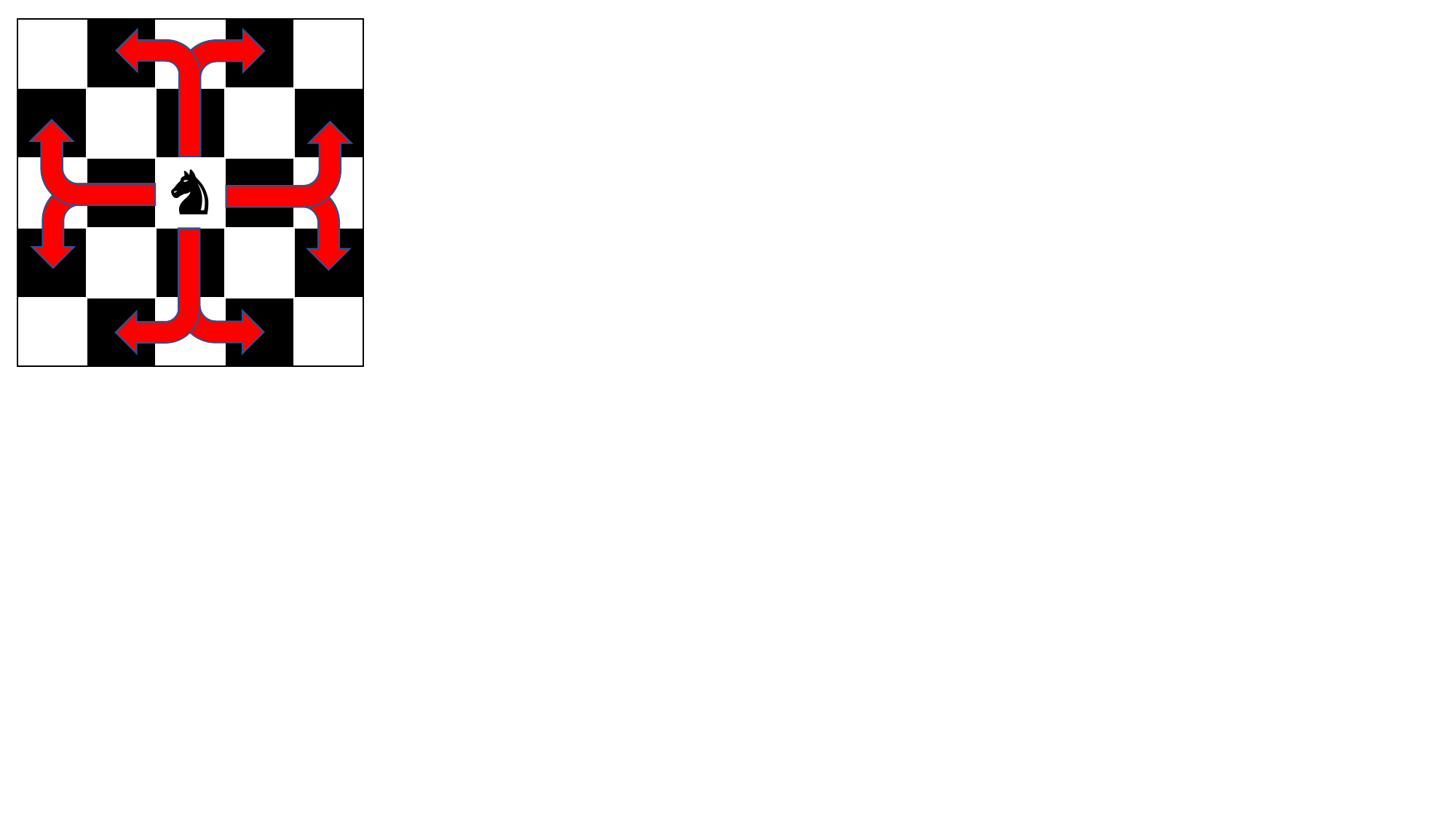}
    \caption{Possible knight moves from a position on a 4 $\times$ 4 chess board.}
    \label{fig:KnightDiagram}
\end{figure}
}

A constraint model of the Knight's Tour is presented in Figure \ref{fig:KnightsTourModel}. We will refer to this as the {\em sequence} model to distinguish it from an alternative presented in Section \ref{sec:KnightsTour}. The model in the figure is a formulation of the problem from the 2008 CSP solver competition\footnote{Archived as of 28 July 2009 at \url{https://web.archive.org/web/20090728111748/http://www.cril.univ-artois.fr/~lecoutre/research/benchmarks/benchmarks.html}}, investigated experimentally by \shortciteA{Rendl2009:automatically} using the available arithmetic and logical operators in the language to capture the legal moves of the knight. However, its performance in terms of constraint propagation is weak, as we will discuss below.

The location of the knight is encoded as a single integer (\(nx+y\)) where \((x,y)\) are the coordinates of the knight on the board (from 0). The \textit{sequence} model simply has a one-dimensional matrix of variables (\code{tour}), with  \code{tour[i]} representing the location of the knight at timestep $i$. The constraints enforce that initially the knight is at the given location, the moves are all different, and each adjacent pair \code{tour[i]} and \code{tour[i+1]} corresponds to a knight's move. As presented in the figure, the knight's move constraint contains two location variables and is naturally expressed using integer division and modulo to obtain the \(x\) and \(y\) coordinates. The constraint states that the absolute difference in the \(y\) coordinates is 1, and for \(x\) coordinates is 2, or vice versa.

\begin{figure}
    \centering
\begin{lstlisting}[language=eprime]
given n: int
given startCol,startRow : int(0..n-1)
find tour : matrix indexed by [int(0..n*n-1)] of int(0..n*n-1)
such that
allDiff(tour),
tour[0] = startCol + (startRow)*n,
forAll i : int(0..n*n-2) .
    ((|tour[i]%n-tour[i+1]%n| = 1) /\ (|tour[i]/n-tour[i+1]/n| = 2)) \/
    ((|tour[i]%n-tour[i+1]%n| = 2) /\ (|tour[i]/n-tour[i+1]/n| = 1))
\end{lstlisting}
\caption{The {\em sequence} model of the Knights Tour in \eprime. The parameters to the model (board size $n$ and the starting square for the knight) are introduced with the keyword {\tt given}. The decision variables (the single-dimensional matrix {\tt tour}) are introduced with the keyword {\tt find}.}
\label{fig:KnightsTourModel}
\end{figure}

The knight's move constraint is a relatively complex expression: a disjunction of conjunctions of reified arithmetic expressions, in which the absolute value of the difference between scaled adjacent {\tt tour} variables is compared with a constant. Typically this constraint will propagate poorly for two different reasons. Firstly, some solvers implement modulo and absolute value poorly. Secondly, the constraint as a whole will propagate poorly, as most solvers will wait until the value of one side of the disjunct is known before propagating the other. 

There are tradeoffs involved in choosing subproblems to tabulate. Tabulating just the expression \code{(|tour[i]\%n - tour[i+1]\%n| = 1)} may well reduce search, but tabulating the entirety of the expression between two adjacent \code{tour} variables in a single step will produce a table of a similar size (as only two variables are involved), while reducing search much more. We could consider tabulating multiple adjacent moves as a single table --- while this may reduce search further, the resulting tables would grow rapidly as a function of the number of variables involved. As we will see, the tabulation opportunities present in this na{\"i}ve model can be identified and exploited automatically by the methods presented in this paper. 

In both of these motivating examples, solver performance is improved by tens or hundreds of times (depending on the solver type) completely automatically by TabID, without the need to apply constraint modelling expertise.

\subsection{Organisation}

The rest of the paper is organised as follows. In \Cref{sec:ast} we review necessary background for the paper. In particular, we provide details of our modelling language, tools, and the abstract syntax tree data structure which our algorithms manipulate.
We then discuss our approach to finding promising parts of the problem where tabulation could be applied, in \Cref{sec:identifying}. Here we cover each of our four heuristics and how they are adapted to different situations that occur in the abstract syntax tree.
In \Cref{sec:tabulation} we describe the tabulation algorithm, its progress checks and work limits, and the caching system. 

We then evaluate our system:
in Section~\ref{sec:eval-part1} in terms of how well the heuristics framework identifies promising subproblems;
in Section~\ref{sec:experimental-eval} in terms of how overall performance is improved compared to the models without tabulation;
in Section~\ref{sec:eval-other} with respect to how well our system performs on a large set of models where tabulation might not a priori be expected to improve performance;
and finally in Section~\ref{sec:scalability} we consider how tabulation scales as we allow the arity of generated tables to increase.
We discuss related work in Section~\ref{sec:related} and conclude in Section~\ref{sec:conc}.


\section{Background}\label{sec:ast}

This section gives the necessary background on constraint propagation, the constraint modelling language we employ in this paper, and the \savilerow tool in which we situate TabID.

\subsection{Constraint Problems, Consistency, and Constraint Propagation} 

A constraint satisfaction problem (CSP) consists of a set of variables, each with a finite domain of values, and a set of constraints that specify the allowed values for given subsets of variables. A solution to a CSP is an assignment of values to the variables that satisfies all of the constraints. To find such solutions, constraint solvers typically combine a search through the space of partial assignments with constraint propagation, a form of deduction that helps to reduce the search required. A constrained optimisation problem (COP) is a CSP with an objective function that must be maximised or minimised.

Constraint propagation usually operates by establishing a {\em consistency} property on the constraints and variables. Generalised arc consistency is a common, powerful consistency property used in this paper, which we define in what follows. The {\em scope} of a constraint $c$, named $\mathrm{scope}(c)$, is the set of variables that $c$ constrains. A \textit{literal} is a decision variable-value pair (written $x \mapsto v$). A literal $x \mapsto v$ is {\em valid} iff $v$ is in the domain of decision variable $x$. A \textit{support} of constraint $c$ is a set of valid literals containing exactly one literal for each variable in $\mathrm{scope}(c)$, such that $c$ is satisfied by the assignment represented by these literals.
A constraint $c$ is Generalised Arc Consistent (GAC)~\cite{DBLP:reference/fai/Bessiere06} if and only if there exists a support for every valid literal of every variable in $\mathrm{scope}(c)$. GAC is established by identifying all literals $x \mapsto v$ for which no support exists and removing $v$ from the domain of $x$. 

There exist efficient constraint propagation algorithms to establish GAC on a table constraint, such as Compact Table \shortcite{demeulenaere2016compact,ingmar2018making} where tuples are bit-packed, Tries \fullcite{Gent2007:data}, and Simple Tabular Reduction \cite{Lecoutre2011:str2}. Table constraints are widely available in constraint solvers.

\subsection{Essence Prime}

The importance of modelling is widely recognised in constraint programming (CP) as well as the related fields of propositional satisfiability (SAT) and integer linear programming (ILP). In CP several constraint modelling languages have been developed, including OPL~\cite{opl-book}, MiniZinc~\cite{nethercote2007minizinc}, and \eprime~\cite{sr-manual} in order to aid in the statement of constraint models and abstract away from the details of particular constraint solvers. Herein, we focus on \eprime, which is comparable with OPL and MiniZinc.

\eprime provides the facility to model parameterised {\em classes} of problems, where an individual problem instance is specified by giving values for the class parameters, for example the integers \code{n}, \code{startCol} and \code{startRow} in Figure \ref{fig:KnightsTourModel}. The language supports Boolean and integer finite-domain decision variables, both singly and collected into multidimensional matrices, such as the one-dimensional matrix of integers \code{tour} in Figure~\ref{fig:KnightsTourModel}. Constraints over these variables are expressed via arithmetic and logical expressions, as can also be seen in the figure. Quantification and comprehension enable the concise statement of such expressions. \eprime supports a number of {\em global} constraints \cite{handbook-cp} that capture common patterns in constraint modelling, including  the all-different constraint present in the figure \cite{Regin1994:filtering}, Global Cardinality Constraint (GCC) \cite{Regin1996:generalized}, and the table constraint that is the focus of this paper.

\subsection{Savile Row, Tailoring, and the Abstract Syntax Tree}\label{sec:intro-ast}

We investigate TabID as a component of the constraint model reformulation tool \savilerow~\cite{sr-journal-17}. 
\savilerow is essentially a multi-pass term rewriting system. It represents a model internally using several abstract syntax trees (ASTs) representing the constraints, the objective function, the domain of each decision variable and parameter (or matrix thereof), and other statements in the model. The parser reads a model in the \eprime language, along with a parameter file giving a value to each of the problem class parameters. There are several backends targeting different solvers, including mature backends for CP and SAT solvers and a prototype ILP backend. The system has a number of different passes, some of which are always performed, others are required for specific backends, and others are optional reformulations intended to improve the performance of the solver. TabID is one optional reformulation. 
We refer to the entire process of transforming a model into input for a solver as \textit{tailoring} the model. 
The early steps of tailoring prior to TabID are as follows:

\begin{itemize}
\item Problem class parameters and other constants (defined by \code{letting} statements) are substituted into the model;
\item All quantifiers and matrix comprehensions are unrolled;
\item Matrices of decision variables are replaced with individual decision variables;
\item Multi-dimensional matrix indexing is replaced with single-dimensional indexing of a one-dimensional version of the matrix if required;
\item Global constraints are identified by simple aggregation steps \cite{sr-journal-17}, e.g.\ collecting a clique of not-equal constraints into a single all-different constraint.
\end{itemize}

In addition, \textit{simplifiers} are applied after each pass to perform partial evaluation and to maintain a normal form. In particular, negation is pushed towards the leaves of the AST (similar to \textit{negation normal form}), double negation is removed, and some operators are rewritten when negated (for example, negated \(=\) is rewritten to \(\neq\)). 
Variable domains are filtered using an external constraint solver prior to tabulation, and any assigned variables are deleted. Details of simplifiers, the normal form, and domain filtering are given by~\citeA{sr-journal-17}. 

\begin{figure}[tbhp]
\begin{center}
\includegraphics[width=0.4\textwidth]{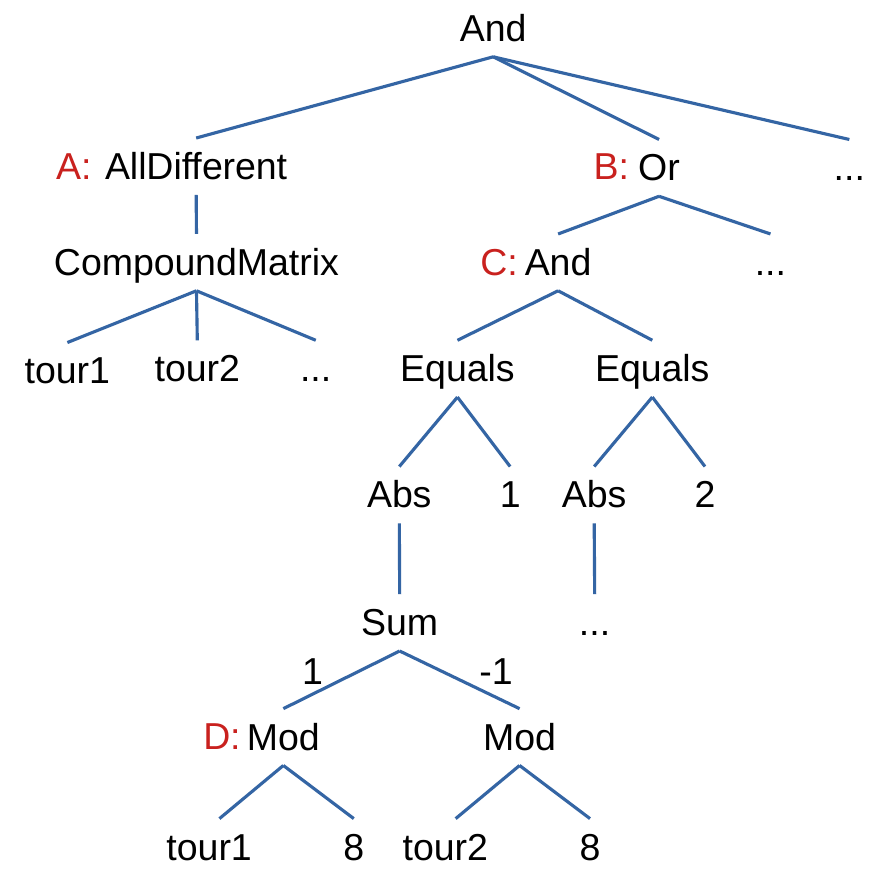}
\end{center}
\caption{\label{fig:ast}Part of the AST representing the constraints of the \textit{sequence} model of Knight's Tour (\Cref{fig:KnightsTourModel}) where \(n=8\), showing the \code{allDiff} constraint and a part of one of the knight's move constraints linking two adjacent \code{tour} variables. Four AST nodes are labelled {\color{red}A} to {\color{red}D}.}
\end{figure}

As an example, consider the Knight's Tour problem and its \textit{sequence} model (shown in \Cref{fig:KnightsTourModel}), with an \(8\times 8\) board. Part of the AST for this model is shown in \Cref{fig:ast}. The \code{tour} matrix has been replaced with individual variables \code{tour0}, \code{tour1}, etc. The \code{forAll} quantifier of \Cref{fig:KnightsTourModel} has been unrolled to create \(n^2-1\) knight's move constraints, each containing two adjacent \code{tour} variables. A fragment of one of the knight's move constraints is shown in the example AST. Also, the variable \code{tour0} has been deleted because it was assigned, and its assigned value has been removed from the other variable domains.

For TabID we consider Boolean and integer expressions, for example those represented by nodes A, B, C, and D in \Cref{fig:ast} (where A, B, and C are Boolean and D is integer). We distinguish between \textit{top-level constraints} and other Boolean expressions. Top-level constraints are Boolean expressions directly beneath the top And node (e.g.\ nodes A and B in \Cref{fig:ast}). We refer to other Boolean expressions as \textit{nested} (e.g.\ node C). Integer expressions are always nested, they cannot be directly contained in the top-level conjunction. 
In terms of the AST a table constraint is a Boolean expression with two arguments: a one-dimensional matrix of decision variables, and a two-dimensional matrix representing a list of satisfying tuples. 
Following the optional TabID pass, some further steps are required to tailor the model for the target solver. We use the default settings of \savilerow which can be summarised as follows, with further details given by~\citeA{sr-journal-17}:

\begin{itemize}
\item Decomposition of constraint types that the target solver does not support;
\item Common subexpression elimination (Active CSE) which factors out identical (or semantically equivalent) expressions replacing them with a new decision variable (for example, in the Knight's Tour \textit{sequence} model in \Cref{fig:KnightsTourModel} there are identical absolute value, division, and modulo expressions that would all be factored out by Active CSE);
\item General flattening to extract nested expressions where the nesting is not allowed by the target solver (for example, in \Cref{fig:KnightsTourModel} the absolute value operator contains a sum which (if not already extracted by CSE) would be extracted and replaced with a new decision variable).
\end{itemize}

In addition the SAT backend has a final encoding step where each integer variable is encoded using \textit{order}, \textit{direct}, or both as required for the constraints containing the variable. The remaining constraints (such as linear, element, and table constraints) are encoded to CNF. Others such as all-different and GCC are decomposed before reaching the SAT backend. Further details of the SAT encoding are given in the \savilerow manual \cite{sr-manual}.


\section{TabID: Identifying Promising Subproblems for Tabulation}
\label{sec:identifying}

We have designed four heuristics for TabID to identify cases where expert modellers might experiment with tabulation to improve the performance of a model. The heuristics operate on the AST, identifying AST nodes that are candidates for tabulation. The heuristics are applied somewhat differently for top-level constraints, nested Boolean expressions, and integer expressions, and we describe these cases in \Cref{sec:heur-toplevel} below. First we describe the types of changes tabulation can make to the AST. 

Note that the heuristics to identify promising subproblems do not take account of the size of the resulting table, nor the work required to generate it. The heuristics can and frequently do identify candidates that would be impractical to tabulate. Progress checks (described in \Cref{sec:tabulation}) allow tabulation to be abandoned early and play an important role in avoiding overhead. 

\Cref{fig:overview} provides an overview of the system, showing how the set of heuristics are combined with other components described in \Cref{sec:tabulation}.

\begin{figure}[tb]
  \centering
  \footnotesize
  \tikzset{%
    inout/.style={draw=ourcolsLineGray,rectangle,fill=white,
      minimum width=2em,minimum height=15pt,text width=12em,text centered},
    stafin/.style={inout,rounded corners,thick},
    proc/.style={draw=none,rectangle,fill=ourcolsLightGray, minimum width=2em, minimum
      height=15pt,text width=12em,text centered},
    srblock/.style={draw=ourcolsLineGray,dashed,thick,fill=none,inner sep=3ex},
    tabidblock/.style={draw=ourcolsHighlight,dashed,very thick,inner sep=2ex},
    arrow/.style = {thick,->,>=stealth},
    edlab/.style = {right,text width=18em}
  }
  \begin{tikzpicture}[node distance=6ex]
    \node (inst) [stafin,text width=24em] {problem instance (\texttt{.eprime} model, \texttt{.param} file)};
    \node (unroll) [proc,below=of inst] {do initial tailoring (see \Cref{sec:intro-ast})};
    \node (idheur) [proc,below=of unroll] {identify promising subtrees using heuristics};
    \node (tabulate) [proc,below=of idheur,yshift=-2ex] {tabulate};
    \node (cache) [proc,left=of tabulate,xshift=-4ex,text width=4em] {cache};
    \node (limits) [proc,right=of tabulate,xshift=4ex,text width=8em] {progress checks,\\work limits,\\SAT size limit};
    \node (target) [proc,below=of tabulate,yshift=-6ex] {re-formulate for target solver};
    \node (solve) [proc,below=of target] {external solver solves};
    \node (intres) [proc,below=of solve] {interpret solver output};
    \node (result) [stafin,below=of intres,text width=16em] {solution, UNSAT, or TIMEOUT};
    \node (tabid) [tabidblock,fit=(idheur) (tabulate) (limits) (cache)] {};
    \node [below left,text width=6em,align=right] at (tabid.north east) {TabID};
    \node (sr) [srblock,fit=(unroll) (tabid) (intres)] {};
    \node [below left,text width=8em,align=right] at (sr.north east) {\savilerow};
    \path
    (inst) edge [arrow] (unroll)
    (unroll) edge [arrow] node [edlab,near start] {initial AST} (idheur)
    (idheur) edge [arrow] node [edlab,text width=8em] {AST candidate expressions} (tabulate)
    (tabulate) edge [arrow] node [edlab,text width=12em,near end] {AST with some expressions replaced by tables} (target)
    (target) edge [arrow] node [edlab] {solver-specific problem file} (solve)
    (solve) edge [arrow] node [edlab] {solver output or timeout} (intres)
    (intres) edge [arrow] (result)
    ;
    \draw ([yshift=.5ex] cache.east) edge [->,>=stealth,bend left=30] ([yshift=.5ex] tabulate.west);
    \draw ([yshift=-.5ex] tabulate.west) edge [->,>=stealth,bend left=30] ([yshift=-.5ex] cache.east);
    \draw ([yshift=.5ex] tabulate.east) edge [->,>=stealth,bend left=30] ([yshift=.5ex] limits.west);
    \draw ([yshift=-.5ex] limits.west) edge [->,>=stealth,bend left=30] ([yshift=-.5ex] tabulate.east);
  \end{tikzpicture}
  \caption{An overview of the TabID process as part of the constraint solving pipeline in \savilerow.
  The white boxes with solid borders represent data; the grey boxes represent
  processes.}
  \label{fig:overview}
\end{figure}

\subsection{AST Modifications}\label{sec:mechanics}

\begin{figure}[t]
\begin{center}
\includegraphics[width=0.6\textwidth]{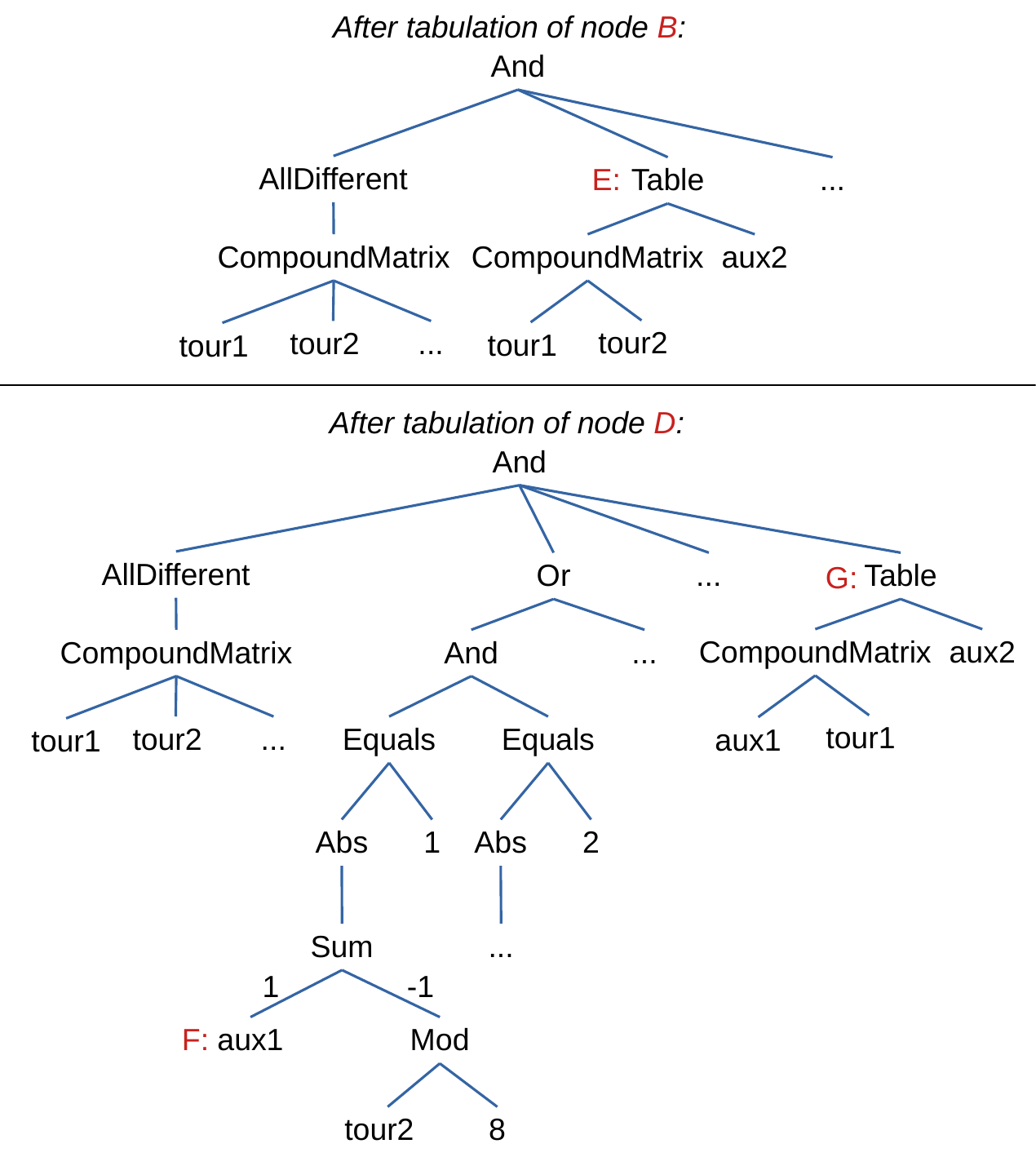}
\end{center}
\caption{\label{fig:ast2}Two examples of tabulation applied to the AST shown in \Cref{fig:ast}. Upper: Node B of \Cref{fig:ast} is replaced with a table constraint labelled E. Both top-level constraints and nested Boolean expressions are directly replaced when they are tabulated. Lower: Node D of \Cref{fig:ast} (an integer expression) is replaced with a new auxiliary variable (F), and a table constraint G is attached to the top-level And node. The scope of the table constraint is the scope of the tabulated expression plus the new variable (\code{aux1} in this case).}
\end{figure}

The AST that TabID acts upon is described in \Cref{sec:intro-ast} with an example in \Cref{fig:ast}. Tabulation modifies the AST in one of two ways depending on the type (integer or Boolean) of the node to be replaced. When the AST node is of type Boolean, the subtree rooted at the node is directly replaced with one table constraint. For example, if node B of \Cref{fig:ast} were to be replaced, the resulting tree would be \Cref{fig:ast2} (upper) in which B has been replaced with node E. Node B is a top-level constraint with variables \code{tour1} and \code{tour2} in scope, and its replacement is a top-level constraint with the same scope. If the node to be replaced is a nested Boolean expression, such as node C in \Cref{fig:ast}, then the replacement table constraint will also be nested. Some solver types do not allow nested table constraints and in this case it would be extracted and replaced with a new Boolean variable, however this occurs in another pass after TabID is complete. 

When the node to be replaced is of type integer, the node is replaced with a new variable and a top-level constraint is created to link the new variable to the variables in scope of the tabulated expression. For example, node D of \Cref{fig:ast} is replaced with new auxiliary variable \code{aux1} in \Cref{fig:ast2} (lower), and a new table constraint G is added to the top-level conjunction with \code{aux1} and \code{tour1} in scope. The new table constraint is generated from an equality between the new variable and the expression to be tabulated, which in this example is \code{aux1 = tour1\%8}.  The domain of the auxiliary variable is generated from the expression using the extended domain filtering method \cite{sr-journal-17} (the default method for all variables introduced by \savilerow). 

Finally, both examples in \Cref{fig:ast2} have an identifier (\code{aux2}) in place of the table of satisfying tuples; this is because matrices of constants are cached to avoid duplication. 

\subsection{Identifying Promising Constraints}\label{sec:heur-toplevel}

First we define the four heuristics as they apply to top-level constraints (i.e.\ constraints in the top-level conjunction). 
The heuristics are as follows, in the order that they are applied:

\begin{description}
\item[Identical Scopes] identifies sets of two or more constraints whose scopes contain the same set of decision variables. 
\item[Duplicate Variables] identifies constraints with at least one variable occurring more than once in the constraint. 
\item[Large AST] identifies a constraint where the number of nodes in the AST is greater than 5 times the number of distinct decision variables in scope. 
\item[Weak Propagation] identifies a constraint \(c_1\) that is likely to propagate weakly (i.e.\ less than GAC), such that there is another constraint \(c_2\) that propagates strongly, with at least one variable in the scope of both \(c_1\) and \(c_2\). The method of estimating whether a constraint will propagate strongly is described in \Cref{sec:strongprop}. Note that only \(c_1\) is identified for tabulation. 
\end{description}

Subproblems containing more than 20 distinct variables are not candidates for tabulation because they would almost certainly be impractical to tabulate. 
Each of the four heuristics is based on a simple rationale regarding either propagation strength or propagation speed of the constraint(s). The constants used in these heuristics were chosen by hand based on preliminary experiments. 

First we consider the Identical Scopes heuristic. It is well known that multiple constraints on the same scope may not propagate strongly together, even if each constraint individually does propagate strongly. The Identical Scopes heuristic is intended to collect such sets of constraints into a single table constraint that may propagate more strongly and also may be faster to propagate. An extreme example would be two contradictory constraints on the same scope (e.g.\ \(x<y\) and \(y<x\)) which would be replaced by a trivially false table constraint. The idea behind Identical Scopes is not new: constraint networks are defined as \textit{normalized} iff no pair of constraints has the same scope~\cite{DBLP:reference/fai/Bessiere06}. 

The Duplicate Variables heuristic identifies constraints that are likely to propagate weakly even when the target solver has a strong propagator for the constraint type. In most cases a GAC propagator will enforce GAC only when there are no duplicate variables. In some cases it is intractable to enforce GAC with duplicate variables. GAC on the Global Cardinality Constraint (GCC) is known to be NP-hard with duplicate variables~\shortcite{Bessiere2007:complexity}, therefore R\'egin's polynomial-time GAC propagator~\cite{Regin1996:generalized} achieves GAC only when there are no duplicate variables. 
The knight's move constraint of the Knight's Tour \textit{sequence} model (\Cref{sec:MotivatingEx}) triggers the Duplicate Variables heuristic: each decision variable in scope is mentioned four times.  In the evaluation we show that tabulating this constraint improves solver performance very substantially. 

The Large AST heuristic identifies constraints that are not compactly represented in the AST. The knight's move constraint of the Knight's Tour \textit{sequence} model also triggers the Large AST heuristic. It has two decision variables in scope, and the AST representation (part of which is illustrated in \Cref{fig:ast}) has 43 nodes. In this case, tabulating the constraint avoids the need to create auxiliary variables during CSE and flattening (\Cref{sec:intro-ast}), and also strengthens propagation of the constraint. 
In general the rationale behind the heuristic is that a table propagator may be more efficient while achieving the same or stronger propagation. 

Finally we consider the Weak Propagation heuristic. It is intended to catch cases where the weak propagation of one constraint is hindering strong propagation of another. 
The knight's move constraint of the Knight's Tour \textit{sequence} model also triggers the Weak Propagation heuristic: the representation of the knight's move constraint for a CP solver is not expected to enforce GAC (because it contains arithmetic such as sum and modulo), and it overlaps with an \code{allDiff} constraint that is expected to enforce GAC. 
To implement the Weak Propagation heuristic we need to estimate which constraint expressions are expected to propagate strongly. \Cref{sec:strongprop} describes how we do this. 

Each of the four heuristics proves to be valuable: each one is triggered on at least one of the problems that we study in the evaluation. The heuristics are applied in the order that they are listed in this section. 

In some cases it can be useful to tabulate an expression within a constraint without tabulating the entire top-level constraint. One example is where the top-level constraint is out of reach of tabulation (i.e.~tabulation would produce an impractically large table or take too long). We have adapted the heuristics in two ways (to identify nested Boolean and integer expressions respectively) as described in \Cref{appn:heur-nested}. The goals remain the same: to strengthen propagation, or to replace an unwieldy expression to improve the efficiency of propagation. The adapted heuristics will be distinguished from top-level heuristics by adding (Nested) or (Integer) to the name (e.g.\ Large AST (Nested), Large AST (Integer)). 

\subsection{GAC Estimate}\label{sec:strongprop}

Given an expression \(e\), the \textit{GAC estimate} is a heuristic to estimate whether the representation of \(e\) for a conventional CP solver will propagate strongly. In cases where CP solvers vary, we use Minion~\shortcite{Gent2006:minion} as a reference. When \(e\) is a top-level constraint, the GAC estimate is simply an estimate of whether the solver will enforce GAC on \(e\). Nested Boolean expressions are treated identically to top-level constraints. When \(e\) is an integer expression, the GAC estimate is applied to the constraint \(a=e\) (where \(a\) is a new auxiliary variable), with the rationale that \(e\) will in many cases be extracted by flattening and replaced with \(a\) (see \Cref{sec:intro-ast}), creating the constraint \(a=e\). 

The definition of the GAC estimate is recursive on the AST representing the expression \(e\). 
To avoid undue complexity the same GAC estimate is used regardless of target solver.
At the leaves of the AST, constants and references to variables are defined to be strong. 
Each type of internal AST node (such as sum or product) has its own rules to define when it is weak or strong. For example, the \code{allDiff} constraint often has a GAC propagator so it is defined to be strong iff all its children are strong. Therefore the constraint \(\mathtt{allDiff}(x_1, x_2, x_3)\) is strong. 
A sum is defined to be strong when all its children are strong, and for each child \(c\) the interval of possible values \([a,b]\) of \(c\) satisfies \(b-a\leq 1\). The reason is that sums are usually implemented with bound consistency propagators that are in general weaker than GAC but equivalent to GAC in this specific case. 
The constraint \(\mathtt{allDiff}(x_1-x_2, x_3-x_4, x_5-x_6)\) on integer variables \(x_1,\ldots,x_6 \in \{1\ldots 3\}\) is therefore defined to be weak. Its representation for a CP solver is unlikely to enforce GAC on the variables \(x_1,\ldots,x_6\) even when \code{allDiff} has a GAC propagator.

\section{TabID: Tabulation, Caching, Progress Checks, and Limits}
\label{sec:tabulation}

Having identified promising candidates, the next step is to perform tabulation efficiently and with appropriate work limits to avoid impractically large tables and long tabulation times. 
The method to generate tables is a straightforward depth-first search and is described in \Cref{sub:gentabs}. For efficiency we have implemented a cache to avoid repeated generation of identical tables for expressions that are semantically equivalent (up to renaming of decision variables). The cache relies on a normal form for expressions, and both the cache and the normal form are described in \Cref{sec:caching}.
Finally, in \Cref{sub:work-limits,sub:satsizelimit} we describe work limits and progress checks applied to tabulation. \Cref{fig:overview} provides an overview showing how the various components of TabID fit together. 

\subsection{Caching}\label{sec:caching}

We use caches to avoid generating identical tables for constraints that are semantically equivalent (up to renaming of decision variables). 
To store or retrieve a table for an expression \(e\), we first place \(e\) into a normal form as follows. First the expression is simplified and normalised as described in \Cref{sec:intro-ast}.
Then all associative and commutative \(k\)-ary expressions (such as sums) and commutative binary operators (e.g.\ \(=\)) within \(e\) are sorted. Alphabetical order is used because it will group together references to the same matrix (all else being equal) and place references to different matrices in a consistent order regardless of the indices. 
The normal form is not only used for accessing the caches. It is also applied before generating a table for an expression \(e\) to ensure that tables are generated with columns in the correct order for storing in the cache. 

After applying the normal form, the expression is traversed in depth-first, left-first order to collect a sequence of decision variables (without duplication), and the variables in the sequence are then renamed to a canonical sequence of names to create \(e'\). Thus the actual variable names in \(e\) do not affect \(e'\), only their relative positions. \(e'\) and the variable domains together are used as a key to store and retrieve tables in the caches. 

As an example, the \(e'\) expression for the Knight's Tour move constraint from \Cref{fig:KnightsTourModel} with \(n=8\) is shown below. 

\begin{lstlisting}[language=eprime]
((1=|((x0%8) - (x1%8))|) /\ (2=|((x0/8) - (x1/8))|)) \/ 
((1=|((x0/8) - (x1/8))|) /\ (2=|((x0%8) - (x1%8))|))
\end{lstlisting}

There are two in-memory caches: the first contains tables, and the second stores cases where tabulation failed because it failed a progress check or reached the node limit. 
The in-memory caches do not persist after one tailoring process on one problem instance. 
We have also implemented a persistent filesystem cache but this is disabled for the evaluation because it would cause timings to change depending on the order of processes. 

\subsection{Progress Checks and Work Limits}\label{sub:work-limits}

In some cases a heuristic will identify a constraint that is too large to be tabulated. Simple work limits (such as those applied in our earlier work~\cite{cp-tabulation-2018} where we limited the depth-first search to generate at most 10,000 tuples, and to fail and backtrack at most 100,000 times) are not ideal because time can be wasted attempting to tabulate constraints that are far beyond reach. 
As an alternative we propose progress checks where the progress of the algorithm through the assignment space is compared to the total size of the space, and if the algorithm seems to be making insufficient progress then the search is stopped early. The depth-first search algorithm progresses through the assignment space in lexicographic order, making it straightforward to calculate the number of total assignments explored so far from the current (partial) assignment. 

Suppose we have a constraint on variables \(x_1,\ldots,x_r\) with domains \(D_1,\ldots,D_r\), and we reached a partial assignment setting variables \(\langle x_1, \ldots x_k\rangle\) to values \(\langle v_1,\ldots,v_k\rangle\) where \(k\leq r\). The partial assignment is completed by filling in the minimum value of the domain for each unassigned variable: 
\[\tau = \langle v_1,\ldots,v_k, \mathrm{min}(D_{k+1}),\ldots, \mathrm{min}(D_r) \rangle\] 
The formulas below assume that each domain is a single contiguous range of integers \(D_i=\{0\ldots \mathrm{max}(D_i)\}\). The implementation has an additional step to map domain values into a single contiguous range. 
The last assignment number \(A\) and current position \(C\) are:

\[A=\sum_{i=1}^r \left[ \mathrm{max}(D_i) \times \prod_{j=i+1}^r |D_j| \right] \]
\[C=\sum_{i=1}^r \left[ \tau_i \times \prod_{j=i+1}^r |D_j| \right] \]

The progress check uses the current node count (\textit{nodeCount}) as well as the parameter \textit{nodeLimit}. It compares the progress made so far to the proportion of the search node limit that has been used so far, effectively using a linear extrapolation to estimate whether the search will complete within the node limit. The search is abandoned if:

\[ \frac{C}{A}  < \frac{\mathit{nodeCount}}{\mathit{nodeLimit}} \]

The term \(\frac{C}{A}\) represents the progress made so far through the search space, and \(\frac{\mathit{nodeCount}}{\mathit{nodeLimit}}\) is the proportion of the search node budget used so far. The progress checks are pessimistic: if a search is expected to slightly exceed the node limit it is abandoned, even though the search algorithm is unlikely to progress through the assignment space at a constant rate. A more optimistic strategy could be obtained by adjusting the formula above (e.g.\ by multiplying the left-hand side by an additional parameter that is \(>1\)), or instead a stochastic estimation procedure could be used~\cite{Knuth1975:estimating}. 
Progress checks are carried out after 1000 and 10,000 nodes, and then after every 10,000 nodes. In addition to the progress checks, the search is terminated if it reaches \textit{nodeLimit} nodes. 

One further limit is applied when tabulating nested Boolean expressions. For solvers or encoding backends that do not support reified or nested table constraints, we limit \(A\) to be no more than \textit{nodeLimit}. The reason is that a nested Boolean expression would be replaced with a nested table constraint by tabulation (as in \Cref{sec:mechanics}) which would then be replaced with a new Boolean variable and a top-level reified table constraint. If the solver does not support reified table constraints, it is converted to a conventional table constraint with exactly \(A\) satisfying tuples. This final step is problematic if \(A\) is large, therefore we limit \(A\). 

\subsection{SAT Size Limit}\label{sub:satsizelimit}

Encoding to SAT and applying a state-of-the-art SAT solver can be very effective for some problem classes. \savilerow has a SAT backend (briefly described in \Cref{sec:intro-ast}) that includes encodings or decompositions for all constraints in \eprime, and with a choice of two encodings for table constraints. Applying TabID prior to SAT encoding can improve solver performance on amenable problem classes, as we will show in \Cref{sec:experimental-eval}. However, in some cases applying TabID can dramatically increase the size of the SAT formula and as a consequence reduce solver performance.

The \textit{SAT size limit} is an optional limit in TabID that prevents large increases in the SAT encoding size (measured by the number of clauses).  Given a candidate Boolean expression \(e\), the SAT size limit first estimates the encoding size of \(e\), then limits the encoding size of the generated table constraint to be no more than 2 times larger than that of \(e\). A small increase is allowed because a table constraint encoding may have better properties than other encodings (for example, unit propagation simulating GAC). 

To estimate the encoding size of \(e\), a CSP instance \(C\) is made containing only \(e\) and decision variables in the scope of \(e\). \(C\) is tailored for encoding to SAT (as described in \Cref{sec:intro-ast}), creating \(C'\).  \(C'\) is encoded to SAT\footnote{With some efficiency measures, e.g.\ generating the entire encoding of \(C'\) is not always necessary.} and the encoding size is recorded (excluding the encoding of variables in \(C\)). 
Finally, the encoding size of the generated table constraint is limited by bounding the number of tuples generated (if the encoding size is a function of the number of tuples), or simply by encoding the generated table constraint and rejecting it if the encoding is too large. 


\section{Feasibility Evaluation of TabID}\label{sec:eval-part1}

We now come to the first part of our evaluation, to show that 
TabID can identify and tabulate promising subproblems of a wide range of models. Here we are evaluating TabID's set of heuristics together with the progress checks and work limits, to show that they are of use in successfully tabulating automatically.  For this feasibility evaluation, we do not consider the effectiveness of tabulations: 
 in \Cref{sec:experimental-eval}, we will show that in many cases applying TabID strongly improves the total time to tailor and solve an instance (including time taken to identify candidates and perform tabulation). 

This feasibility evaluation is divided into two.
We first examine, in Section~\ref{subsec:eval-part1-baseline}, what we call
`Baseline Problems' where we have identified examples of tabulation in the literature.
 Tabulation (whether performed manually or with tool support) is a well-established technique, and for some problem classes and models there are examples in the literature of subproblems that can profitably be tabulated. For these models we compare with the literature, in addition to comparing the original model to the version after applying TabID.
 We then examine, in Section~\ref{sec:exp2}, seven new case studies where tabulation has not been previously identified in the literature but where TabID was able to tabulate automatically.
In Section~\ref{sub:feas-results-summary} we summarise the results of our study and how each heuristic performs. 

Following some preliminary experiments, we have set the parameter \textit{nodeLimit} to 100,000 for our feasibility evaluations. Also, we chose not to apply the SAT size limit. The effect of the node limit can be seen (for example) with the Killer Sudoku problem where some constraints of arity 5 (on variables with domain size 16) fail a progress check but others are successfully tabulated. 
The base models (with parameter files) are publicly available online \shortcite{software-release-tab} alongside a version of \savilerow that implements TabID. 

\subsection{Baseline Problems}\label{subsec:eval-part1-baseline}

The set of baseline problems consists of four problems presented by~\citeA{Dekker2017:autotabling}, and two others: Sports Scheduling Completion and Maximum Density Still Life. In the first four cases we show that TabID can automatically identify and tabulate the same subproblems that Dekker et al.\ identified by hand and found to be useful, as well as finding some other opportunities for tabulation. Similarly, for Sports Scheduling Completion, TabID identifies and tabulates the same constraint as in the literature. For Maximum Density Still Life, there is no exact equivalent in the literature but TabID tabulates the ``easy formulation'' constraint from the model of~\citeA[Section 1.1]{Bosch2004:constraint}.


\subsubsection{Black Hole}

Black Hole was introduced as a variant of patience in Section \ref{sec:MotivatingEx}, along with a constraint model of the problem. To recap, the model has two matrices of variables: \code{blackHole}, the sequence of cards played into the black hole; and its inverse \code{cardSequence} (the index of each card in \code{blackHole}). We post the adjacency constraint on each pair of adjacent variables in \code{blackHole}. Less-than constraints on \code{cardSequence} ensure that the cards in a fan are not played out of order. Both matrices have an \code{allDiff} constraint, and they are linked by channelling constraints. An instance of Black Hole is a permutation of the 52 cards. We experimented with the 102 randomly-generated instances from CSPLib~\cite{csplib:prob081}. 

All adjacency constraints trigger the Weak Propagation heuristic because they overlap with the \code{allDiff} on the \code{blackHole} matrix. In addition the Identical Scopes (Nested) heuristic is triggered by a small number of equalities (no more than 9)  within the channelling constraints. No other constraint triggers any heuristic, so our set of candidates is very similar to those identified by hand \fullcite{Gent2007:search}. All candidates are successfully tabulated. 

\comment{
IS(Nested) doesn't even change node count with PN-1 and Gecode. 
}


\subsubsection{Block Party Metacube Problem}\label{sec:bpmp-eval1}

The Block Party Metacube Problem is a puzzle in which eight small cubes are arranged into a larger \emph{metacube}, such that the visible faces on each of the six sides of the metacube form a `party'.
Each small cube has a symbol at each corner of each of its faces (24 symbols per cube in total), and each symbol has three attributes, with each attribute in turn taking one of four values.
To form a valid party (the \textit{party constraint}), the four small cubes forming a visible face of the large cube must be arranged so that the four symbols in the middle of the visible face are either all different, or all the same, for each of the three attributes. 

The model we use is closely based on~\citeA{Dekker2017:autotabling} and we use the same set of instances. The model has two matrices of decision variables, \code{cubeAt} and \code{symAt}. 
Matrix \code{cubeAt} encodes a permutation of the 8 cubes (numbered 1 to 8), representing the relative locations of the cubes in the metacube.
Further, \code{symAt} represents for each of the 4 symbol positions located in the middle of each of the 6 faces of the metacube (24 symbol positions) which symbol is visible in that position.
A hand-computed matrix \code{pp} encodes how the 24 positions in which symbols are placed on a cube occur together at corners of a cube. A set of channelling constraints link each \code{cubeAt} variable to three \code{symAt} variables. 

There are some notable differences between our model and Dekker et al. They introduce a variable for each attribute at each of the 24 symbol positions, whereas in our model the expression for the attribute (one of the following: \code{symAt[i]/16}, \code{symAt[i]\%4}, or \code{(symAt[i]\%16)/4} with constant \code{i}) is used wherever it is needed. Where necessary \savilerow will introduce a variable for an attribute expression during tailoring. Also, Dekker et al.\ introduced a local \textit{rotation} variable for each small cube in their non-tabulated model. The rotation variables are not present in their tabulated model. \eprime does not have local decision variables so we used an existential quantifier in place of each rotation variable. 

Dekker et al.\ tabulated the 8 channelling constraints linking cubes and symbols. 
The Duplicate Variables, Large AST, and Weak Propagation heuristics all identify the same set of channelling constraints (with arity 4) and they are all successfully tabulated. 
The Weak Propagation (Nested), Large AST (Nested), or Weak Propagation (Integer) heuristics identify the attribute expressions  (e.g.\ \code{symAt[i]/16}) or an equality of two attribute expressions, contained in the party constraints. All such expressions are successfully tabulated (and all have arity 2). Larger nested sub-expressions of the party constraints are identified by Identical Scopes (Nested), but almost all fail the progress check at 1,000 nodes. At most 4 are tabulated in any given instance.

\comment{
Checked+updated 22/5/24

Verified that it tabulates the 8 channelling constraints. 

1176 -- 4 sub-expressions of party constraints are tabbed in 1176. Just within limit. 

1266 -- 2

2254 -- 0

3237 -- 2

3345 -- 0

412 -- 4

4629 -- 0

7059 -- 2

7324 -- 2

7326 -- 0

7444 -- 4

7661 -- 0

7750 -- 4

8848 -- 2

Sampled 1176, 2254 and 3237 to compare 

1176 -- smaller node count with TabID vs just tab channelling cts, but same speed Minion
2254 -- smaller node count but 0.5s slower with TabID vs tab channelling cts.
3237 -- smaller node count but 20\% slower with TabID vs tab channelling, with Minion. 

Reran these with gecode:
1176 -- 10s to 7s with TabID in addition to tab channelling cts. Half nodes. 
2254 -- similar times, fewer nodes, 
3237 -- 19s to 16s with TabID, ~half nodes. 
}


\subsubsection{Handball Tournament Scheduling}\label{sec:hts-eval1}

The Handball Tournament Scheduling problem is to schedule matches of a tournament, while respecting the rules governing the tournament, and minimising a cost function related to the availability of venues. We use the model and the same 20 instances of~\citeA{Dekker2017:autotabling}, which is simplified from the full model~\shortcite{Larson2014:integrated} by omitting some constraints. 
The problem has 14 teams (in two divisions of 7), and briefly it requires constructing one round-robin tournament for each division (in periods 1-7), followed by a round-robin tournament for all teams (periods 8-20) then its mirror image (periods 21-33). 
Constraints are either structural (such as balancing home and away games) or seasonal (such as respecting venue unavailability, given as a parameter). 
The main sets of variables represent: the \textit{home-away pattern} (\code{HAP}) indexed by row and period, with values \emph{home}, \emph{away}, or \emph{bye} (i.e.\ does not play); the \code{break} period for each row of \code{HAP}, when the alternating home-away pattern is broken by the team playing at home twice or away twice in sequence; the \code{contestant} (opponent) for each entry in \code{HAP}; the team assigned to each row of \code{HAP}, named \code{teamof}; and the cost of each row (\code{rowcost}), which contributes to the objective. 

\comment{
NO LONGER TRUE SINCE HAP <-> BREAK CHANNELLING IS IN A TABLE. 
The original model contains \emph{regular} constraints (where the meaning of the constraint is defined by a deterministic finite-state automaton) and we use a manual decomposition of \code{regular} (via its finite-state automaton) because \savilerow does not currently implement \code{regular} constraints. The \code{regular} constraints are applied to the home-away pattern matrix. The decomposition introduces a state variable for each step along the sequence and generates constraints of the following form (where \code{state} are the auxiliary state variables, \code{M} is a constant matrix, \code{t}, \code{d} and \code{k} are constants):

\begin{lstlisting}[language=eprime]
state[t+1] = M[d*state[t] + seq[t] - k]
\end{lstlisting}
}


Dekker et al.\ reformulated the original model in three steps. First, they generated a \code{table} constraint to channel between a row of \code{HAP} and its corresponding \code{break} variable (and reported no significant speed-up from this). Second, they reformulated the row cost constraint for each row of \code{HAP} to be in terms of \code{break}, \code{teamof}, and \code{rowcost} (replacing \code{HAP} variables with \code{break}). Thirdly, they applied tabulation to the row cost constraints (generating arity 3 \code{table} constraints). 

TabID is not capable of reformulating a constraint to change its scope, so we have performed the first two steps manually and we evaluate whether TabID can do the final step automatically. The model presented to TabID closely follows Dekker et al., except that local decision variables declared within the row cost constraint are replaced with an existential quantifier (because \eprime does not have local decision variables). The quantifier unrolls to a disjunction where each disjunct corresponds to one value of the \code{break} variable. 

When TabID is disabled, using a disjunctive version of the row cost constraint would be artificially bad so in this case we use a different formulation that is stated on a row of the \code{HAP} matrix, \code{teamof}, and \code{rowcost}. The formulation of the objective is similar to the baseline (non-tabulated) model of Dekker et al. The rest of the model is identical to the version presented to TabID. 

The Duplicate Variables, Large AST, and Weak Propagation heuristics identify the row cost constraints and all are tabulated. Also, unary constraints on \code{contestant} are identified by Weak Propagation, while Identical Scopes (Nested) identified equalities containing a single \code{contestant} variable. 
Weak Propagation (Integer) triggered for shift expressions used in matrix indexing and linear expressions involving \code{contestant} and \code{HAP}. All candidates mentioned here were tabulated. 

\comment{
Checked 3/6/24

Connecting break[i], teamof[i] and rowcost[i]: DV, LAST, WP

Unary constraints on contestant: WP

More unary constraints on contestant: IS(Nested)
-- ((((-3 + contestant_00008_00003) in int(1,3,5,7,9,11))) /\ (8=contestant_00008_00003))

Shift expression contained in element that looks up in HAP. : WP(Integer)

Linear expression containing contestant and HAP: WP(Integer)

With Gecode: TabID improves on model with only rowcost constraint tabulated. 94s -> 73s on 01.param. 
}


\subsubsection{JP Encoding Problem}\label{sec:jpe-eval1}

The JP Encoding problem was introduced in the MiniZinc Challenge 2014. In brief, the problem is to find the most likely encoding of each byte of a stream of Japanese text where multiple encodings may be mixed. The encodings considered are ASCII, EUC-JP, SJIS, UTF-8, or unknown (this choice incurs a large penalty). Once again our model closely follows that of~\citeA{Dekker2017:autotabling}. We use all 10 instances in the MiniZinc benchmark repository. The instances are from 100 to 1900 bytes in length.  Each byte has four variables: the \code{encoding}, a \code{byte\_status} variable that combines the encoding with the byte's position within a multibyte character, a \code{char\_start} variable indicating whether the byte begins a new multibyte character, and the \code{score} which contributes to the objective.

Dekker et al.\ tabulate three subproblems. The first connects two adjacent \code{status} variables, and the Identical Scopes heuristic triggers on this. The second links \code{status}, \code{encoding}, and \code{char\_start}, and we found that the Identical Scopes heuristic separately links \code{status} to \code{encoding}, and \code{status} to \code{char\_start}. The \code{encoding} and \code{char\_start} variables are both functionally defined by \code{status} so no propagation is lost with two binary table constraints compared to one ternary table. Thirdly Dekker et al.\ tabulate the constraint linking the score to the encoding. The Duplicate Variables and Large AST heuristics trigger on this. 
In summary, the set of subproblems identified by TabID is not identical to Dekker et al., but is exactly equivalent in propagation strength (assuming GAC is enforced on table constraints). All candidates are successfully tabulated (creating binary tables). 

\comment{
Checked 27/5/24

data100:
Adjacent status variabls: IS

Encoding <-> byte status : IS

byte status <-> char start: IS

encoding <-> score ct : dup vars, LAst. 

data1900: same
}


\subsubsection{Maximum Density Still Life}
\label{sec:negative-example}

The Maximum Density Still Life problem, CSPLib problem 32~\cite{csplibprob032-revised}, is to maximise the number of live cells in an $n \times n$ grid in such a way that applying the rules of John Conway's Game of Life would leave the grid unchanged.  Cells outside the grid are assumed to be dead. The rules state that a live cell survives if it has two or three live neighbours, but dies otherwise. A dead cell will only become alive if it has exactly three live neighbours. Maximum Density Still Life has been solved for extremely large grids \cite{Chu2012:complete} using a sophisticated CP model and other techniques. For sufficiently large grids, a provably optimal solution is constructed from a finite set of tiles, thus providing a solution for any \(n\). 

Our focus is not on solving Still Life for large \(n\) but in evaluating tabulation, and so we have simply modelled the problem using an \(n\times n\) matrix \code{g} of Boolean variables (where \textit{true} means live), surrounded by a border of width 2 of cells that are \textit{false}. The rules are implemented with two implication constraints, one setting the cell to \textit{true} if exactly three of the eight neighbours are alive; the other setting the cell to \textit{false} if fewer than 2 or more than 3 neighbours are alive; leaving cells with exactly two live neighbours unconstrained, as they would be unchanged in the next step of the game. The constraints are applied to each cell in the \(n\times n\) matrix and to the inner layer of border cells. We used the instances where \(n\in \{6..15\}\). 
Letting \code{sum(neighbours)} abbreviate the sum of the 8 neighbours of \code{g[i,j]}, 
the constraints for one cell \code{g[i,j]} are written as follows:

\begin{lstlisting}[language=eprime]
sum(neighbours) = 3 -> g[i,j],
(sum(neighbours) > 3 \/ sum(neighbours) < 2) -> !g[i,j]
\end{lstlisting}

For each cell within the \(n\times n\) matrix, the Identical Scopes heuristic identifies the two implication constraints (together) as a candidate for tabulation, and they are successfully tabulated (with arity 9 for cells not touching the edges). For the cells in the inner border (and away from the corners), each cell has 3 unassigned neighbours and \savilerow simplifies the two constraints to a single \(\mathrm{sum}\neq 3\) constraint which is not a candidate for tabulation. When tabulation is disabled, the three occurrences of \code{sum(neighbours)} are removed by common subexpression elimination~\cite{sr-journal-17} and replaced with a new variable. 

Our model is logically equivalent to the ``easy formulation'' constraint model of \citeA[Section 1.1]{Bosch2004:constraint}.
Tabulation seemed promising as an improvement to this model without needing to reformulate the constraints to take into account additional mathematical insights~\cite{Chu2012:complete,Smith2002:dual}. The constraints may seem to be natural candidates for tabulation, but in fact (as we will see in \Cref{sec:exp-baseline}) tabulation provides no benefit, slowing down solving in most cases. We therefore include Maximum Density Still Life as an example of tabulation where the result negatively affects performance. 

\comment{
Checked 27/5/24
}


\subsubsection{Sports Scheduling Completion}

The Sports Scheduling Completion problem is to construct a schedule of \(n(n-1)/2\) games among \(n\) teams (\(n\) must be even), where each team plays every other team once. The schedule is divided into \(n-1\) weeks, in each week there are \(n/2\) periods, and one game is played in each period of each week. No distinction is made between home and away games. Each team plays at most twice in a period, and each team plays exactly once each week~\shortcite{vanhentenryck1999:constraint}.

The schedule is represented explicitly with a matrix \code{schedule[w,p,i]}, indexed by the week \code{w}, period \code{p}, and \code{i} which is 1 or 2 for the two teams in the game. Some symmetry is broken by ordering the two teams in each game in the \code{schedule} matrix. \code{allDiff} constraints are used to ensure each team plays once each week, and a set of \code{gcc} constraints (one per period) ensure each team plays at most twice in each period (with an auxiliary variable \code{card} indicating the number of games played by each team in each period). 
A second matrix \code{game[w,p]} (also indexed by week and period) represents each game with a single integer. The two representations are channelled with the following constraints:

\begin{lstlisting}[language=eprime]
forAll w : WEEKS . forAll p : PERIODS .
    game[w,p] = n*(schedule[w,p,1]-1) + schedule[w,p,2]
\end{lstlisting}

Finally an \code{allDiff}  constraint is posted on the \code{game} matrix to ensure every team plays every other team exactly once. 

In Sports Scheduling Completion, we start with a partial schedule where some of the slots in \code{schedule} are assigned. 10 instances were generated with \(n=12\) and 10 slots assigned a team at random with uniform distribution.  Trivially unsatisfiable instances were filtered out. 

The Weak Propagation heuristic identifies each of the channelling constraints (on 3 variables), and all are tabulated. \citeA{vanhentenryck1999:constraint} manually tabulated the same constraint in their OPL model of Sports Scheduling. Weak Propagation also identifies linear equality constraints over the \code{card} variables for each period (introduced automatically as implied constraints from \code{gcc}), and these are also tabulated. 

\comment{
Checked+updated 27/5/24

channelling : WP

sum caps = constant:  WP
}


\subsection{New Case Studies}\label{sec:exp2}

In this section we present seven case studies that were not featured in the literature to our knowledge. In each case we briefly describe the model and discuss the expressions that trigger the TabID heuristics.


\subsubsection{Accordion Patience}\label{sec:accordion-eval1}

`Accordion' \cite{accordion-rules} is a single-player (patience or solitaire) card game.
The game starts with the chosen cards in a sequence, each element of which we consider as a `pile' of one card. Each move we make consists of moving a pile on top of either the pile immediately to the left, or three to the left (i.e.\ with two piles between the source and destination) such that the top cards in the source and destination piles match by either rank (value of the card, e.g.\ both 7) or suit (clubs, hearts, diamonds, or spades). The result of each move is to reduce the number of piles by 1 and change the top card of the destination pile. The empty space left at the position of the source pile is deleted. The goal is to keep making moves until just one pile remains. 
We consider the `open' variant where the positions of all cards are known before play starts, in a variant studied by \citeA{accordion-stanford} where we play with a randomly chosen subset of $n \leq 52$ cards. A problem instance consists of the subset of cards in play, and their initial positions. We have generated 5 random instances for each value of \(n\in \{11,12,13,14,15,16\}\). 

We model accordion with a matrix called \code{piles} of $(n-1) \times n$ variables with domain $\{0\ldots 51\}$, the element \code{piles[i,j]} representing the top card of the stack in position $j$ in the sequence after move $i$. Only the top card in each stack is represented; others are not relevant. We have $2(n-1)$ decision variables for the $n-1$ moves, expressing which pile is moved to which other pile (named \code{from} and \code{to}).  We also have two variables per move representing the top card of the stack that is moved (\code{fromcard}), and the top card of the stack it is moved onto (\code{tocard}). Frame axioms ensure that unmoved cards are copied from one timestep to the next, and that the unused slots at each timestep fill up with zeroes. A set of constraints link \code{to} with \code{tocard}, and \code{from} with \code{fromcard} by indexing the \code{piles} matrix. Finally the move is implemented by indexing into \code{piles} with \code{from} and \code{to}. 

There are two key constraints on the moves. The first (Move1) is that the move is of either one or three places, written as follows:

\begin{lstlisting}[language=eprime]
forAll t : int(1..n-1) . to[t] = from[t] - 1 \/ to[t] = from[t] - 3
\end{lstlisting}

The other key constraint (Move2) ensures that the top cards of the two piles are of the same rank or suit. This is expressed by stating that the relevant two cards either have the same value modulo 13 or integer-divided by 13, as follows:

\begin{lstlisting}[language=eprime]
forAll t : int(1..n-1) .
  fromcard[t]%13 = tocard[t]%13 \/ fromcard[t]/13 = tocard[t]/13
\end{lstlisting}

All Move1 and Move2 constraints are identified by the Duplicate Variables, Large AST, and Weak Propagation heuristics and are tabulated, creating binary tables. The Weak Propagation (Integer) heuristic identifies expressions of the form \code{x-c} where \code{c} is a constant. These expressions come from indexing into the \code{piles} matrix with \code{from} and \code{to}, and they are tabulated. Identical Scopes identifies a small number of frame axiom constraints which are tabulated. Preliminary experiments showed that almost all benefit came from tabulating Move1 and Move2 constraints. 

\comment{
Checked 28/5/24

Move1 ct: dup vars, last, weakprop

Move2 ct: same

Frame axioms: IS

Shift: WP(Int)

Testing cards_13_01.param with Minion:
TabId: 241s,   tabulate move1+move2 only: 225s (nodes only 1 different)
Same with Gecode:
TabID: 56s, tab only move cts: 55s (same nodes)
}


\subsubsection{Coprime Sets}

Erd\H{o}s and S\'ark\"ozy~\citeyear{Erdos1993:sets} studied a range of problems involving coprime sets.
A pair of numbers $a$ and $b$ are coprime if there is no integer $n>1$ which is a factor of both $a$ and $b$. The Coprime Sets problem of size $k$ is to find the smallest $m$ such that there is a subset of $k$ distinct numbers from $\{m/2\ldots m\}$ that are pairwise coprime.
In our model, the set is represented as a sequence \code{V} of integer variables. Each pair of variables \code{V[i]} and \code{V[j]} has a set of coprime constraints, one for each potential factor in \(\{2\ldots m\}\):

\begin{lstlisting}[language=eprime]
forAll d : int(2..m) . ((V[i]%d != 0) \/ (V[j]%d != 0))
\end{lstlisting}

Adjacent variables in the sequence are ordered with less-than constraints to break symmetry. Also, the lower bound of \(m/2\) is enforced with constraints \code{V[i]>=(V[k]/2)} for each \code{i} from 1 to \code{k-1}. Finally the variable \code{V[k]} is minimised. 

In the experiments, we used the instances where \(k \in \{8\ldots 25\}\). 
For each pair of variables, the Identical Scopes heuristic is triggered by the coprime constraints, a symmetry breaking constraint if one exists, and a lower-bound constraint if one exists. For instances of size \(8\ldots 19\), tabulation is successful for each candidate set of constraints, and all original constraints are replaced with binary tables. 
For the larger instances, Large AST (Nested) and Weak Propagation (Nested) identify sub-expressions of the coprime constraint: \code{(V[i]\%d != 0)}, and these are tabulated. 

\comment{
Checked 28/5/24

15: OK, IS only, all tabulated. 

17: OK

18: OK

19: OK

20: no, only these and a couple of others:

LargeAST,WeakPropagation(Nested)   Attempting tabulation: ((V_00018\%392) != 0)
Tabulated by retrieving from cache.

25: no, only these:

LargeAST,WeakPropagation(Nested)   Attempting tabulation: ((V_00003\%585) != 0)
Tabulated by retrieving from cache.
}


\subsubsection{Killer Sudoku}

Killer Sudoku, CSPLib problem 57~\cite{csplibprob057-revised}, is a popular puzzle similar to the classical Sudoku, where an empty $9\times 9$ grid is filled in with numbers $1\ldots 9$, such that each row, column and the nine 
non-overlapping $3\times 3$ sub-squares take different values. 
In Killer Sudoku there are also clues (which differ between instances). Clues are sets of cells that sum to a given value (and also take different values). We use a straightforward model where each cell of the grid has one decision variable with domain $\{1\ldots 9\}$. The variables of each row, column and non-overlapping $3\times 3$ sub-square are constrained by \code{allDiff}. Traditional $9\times 9$ Killer Sudoku instances are trivial for a constraint solver, so we use an existing model for the \(16\times 16\) case, and an existing set of 100 \(16\times 16\) instances \cite{sr-journal-17}. The instances are all satisfiable but unlike conventional Killer Sudoku puzzles, they may have more than one solution. 

Each clue is a set of cells (from 1 to 5 cells) that are contiguous. For each clue, the model contains both an \code{allDiff} (except for size 1 clues) and a sum equality constraint on the same scope. In a conventional constraint solver (without clause learning), the model propagates poorly and solving times can be poor. It was shown by \citeA{sr-journal-17} that associative-commutative common subexpression elimination (AC-CSE) (when combined with implied sum constraints generated from the \code{allDiff} constraints) can improve solving times substantially by connecting the two constraints on each clue, and also connecting the clues to the rules. The constraints below represent a size 3 clue from one of the instances. 

\begin{lstlisting}[language=eprime]
(field[3,4] + field[4,4] + field[4,5]) = 26,
allDiff([field[3,4], field[4,4], field[4,5]])
\end{lstlisting}

For each clue of size 3 to 5, the two clue constraints are (together) identified as a candidate by the Identical Scopes heuristic. For clues of size 2, the clue \code{allDiff} is removed prior to tabulation because it is subsumed by a larger \code{allDiff}, leaving just the sum equality constraint. The sum equality triggers the Weak Propagation heuristic. Clues of size 2 to 4 are tabulated, but in some cases clues of size 5 cannot be tabulated. For example, the first instance (named \texttt{sol1} in the repository) has 9 clues of size 5, of which 6 were successfully tabulated. The other 3 failed the progress check after 10,000 nodes (as described in \Cref{sub:work-limits}). In each case the arity of the generated table is equal to the size of the clue. 

\comment{
Checked 30/5/24 -- OK
}


\subsubsection{Knight's Tour Problem}\label{sec:knights-eval1}
\label{sec:KnightsTour}

The Knight's Tour Problem was described in Section \ref{sec:MotivatingEx}. Recall that we use the Hamiltonian path version of Knight's Tour, i.e.\ the last square visited is not required to be a knight's move from the first square visited. An instance defines \(n\) and the starting location of the knight. We experiment with instances where \(n\in \{6\ldots 12,15,20,25,30,35\}\) and with two starting locations, \((0,0)\) and \((0,1)\), for 24 instances in total. 

We use two models, and in both the location of the knight is encoded as a single integer (\(nx+y\)) where \((x,y)\) are the coordinates of the knight on the board (from 0). The first model is the \textit{sequence} model introduced in Section \ref{sec:MotivatingEx} (see Figure \ref{fig:KnightsTourModel}), in which we have a one-dimensional matrix of variables (\code{tour}), with  \code{tour[i]} representing the location of the knight at time-step \code{i}. The constraints enforce that initially the knight is at the given location, it never revisits a location (via \code{allDiff}), and each adjacent pair \code{tour[i]} and \code{tour[i+1]} corresponds to a knight's move. The knight's move constraint uses integer division and modulo to obtain the \(x\) and \(y\) coordinates. For convenience we re-cap this constraint here:

\begin{lstlisting}[language=eprime]
((|tour[i]%n - tour[i+1]%n| = 1) /\ (|tour[i]/n - tour[i+1]/n| = 2)) \/
((|tour[i]%n - tour[i+1]%n| = 2) /\ (|tour[i]/n - tour[i+1]/n| = 1))
\end{lstlisting}

The second model (named \textit{successor}) has a matrix of variables \code{next} which indicate the successor of each location. The \code{next} variables are constrained by \code{allDiff}. In addition, it has all variables and constraints of the \textit{sequence} model and a set of channelling constraints connecting \code{next} to \code{tour}, ensuring there are no cycles in \code{next}.\footnote{The standard CP model of a Hamiltonian path problem uses a successor viewpoint and a global constraint (e.g.\ \code{path} in Gecode~\cite{gecode-website-mod}). \savilerow and Minion do not have the path constraint. Instead the \textit{successor} model loosely follows an example in Gecode~\cite{gecode-website-mod} (credited to Gert Smolka) which has successor, predecessor, and \textit{jump} variables that give the index of each location in the sequence. In preliminary experiments our model performed slightly better than the Gert Smolka model when using Minion. The size of both of our models and Gert Smolka's model (as the sum of domain sizes) is \(\Theta(n^4)\). 
} The channelling constraints are as follows: 

\begin{lstlisting}[language=eprime]
forAll i : int(0..tourLength-2). next[tour[i]] = tour[i+1]
\end{lstlisting}

The knight's move constraint triggers the Duplicate Variables, Large AST, and Weak Propagation heuristics. For both models, when \(n\leq 15\), all knight's move constraints are tabulated (creating a binary table). For instances where \(n=20\), the first 15 or 16 are tabulated (where variable domains have been reduced by preprocessing), and the number reduces further to 9 when \(n=35\). For the remaining knight's move constraints, each division and modulo operator is identified by Weak Propagation (Integer), extracted and tabulated (creating a binary table constraint). Each unique division and modulo expression is tabulated once and the auxiliary variable is reused as described in \Cref{sub:nested-int}. When TabID is disabled, identical common subexpression elimination (CSE), which is part of the default configuration~\cite{sr-journal-17}, improves the knight's tour constraint by adding auxiliary variables for the division, modulo, and absolute value expressions. 

Finally, the channelling constraints in the \textit{successor} model are translated to an \code{element} constraint that indexes from 1. If the lower bound of \code{tour[i]} is not 1 (after domain filtering) then a shifted index expression \code{tour[i]+c} is created.\footnote{For instances where the starting location is \((0,1)\) most \code{tour[i]} variables have a lower bound of 0.} The \code{tour[i]+c} expressions are identified by Weak Propagation (Integer) and tabulated for the 8 instances where \(n\leq 15\) (also creating binary tables). When \(n>15\) a small proportion of them are tabulated. 

\comment{
Checked+revised sequence model 31/5/24

Knights (sequence):
n=35a,  does first 9 cts entirely, DuplicateVariables,LargeAST,WeakPropagation

rest:
WeakPropagation(Integer) tabulation: ((tour_00009\%35)=aux9)

n=35b same as above. 

n=20a
first 16 cts done entirely

20b
15 done entirely.

n=15a,b
all done entirely. 

Successor model:
Should be identical for main constraints
Then the shifts tour[i]+c are done by WP(Int)
}


\subsubsection{N-Linked Sequence and Optimal N-Linked Sequence}
\label{sec:nlinked}

This puzzle, proposed by Itay Bavly~\cite{nlinkedseq1}, requires arranging as many as possible out of the first 100 positive integers into a sequence, so that in every pair of adjacent numbers one is a multiple of the other.
The longest possible sequence was found to consist of 77 numbers~\cite{nlinkedseq2}. The question was also asked for 1000 numbers, and in this case a sequence with 418 numbers was constructed.
We use the name \textit{n-linked sequence} as proposed by William Gasarch~\citeyear{Gasarch2019:open},
who also introduced the parameter $n$ for the largest integer.

We consider two versions of the puzzle. The first version is a decision problem in which we ask whether a sequence exists of some given length, where the length \code{len} is a fraction of $n$ chosen to be challenging (close to the unsatisfiability threshold, but still satisfiable). The second version is an optimization problem where we simply seek a longest sequence as in the original problem description. Both versions are modelled with a one-dimensional matrix \code{seq} of variables with domain \(\{1\ldots n\}\) representing the sequence. In the decision version, \code{seq} is of length \code{len}, which is a parameter.  In the optimization version, \code{len} is a variable that is maximised. In both versions the entire matrix \code{seq} is \code{allDiff}. The divisibility constraint for the optimization problem is shown below. In the decision problem, the condition \code{(i<=len)} is omitted. 

\begin{lstlisting}[language=eprime]
(i<=len) -> ((seq[i]%seq[i-1] = 0) \/ (seq[i-1]%seq[i] = 0))
\end{lstlisting}

For the decision version of the problem, we used the 15 instances where \(n\in \{60,70,80\}\) and \(\mathit{len} \in \{n-30, n-25, n-20, n-15, n-10\}\). For the optimization version, we used instances \(n=\{12, 14, 16,\ldots,42\}\).  For the optimization problem, the Duplicate Variables and Weak Propagation heuristics identify each arity 3 divisibility constraint and they are all tabulated. For the decision problem, Duplicate Variables, Large AST, and Weak Propagation heuristics identify each arity 2 divisibility constraint and all are tabulated.

\comment{
Checked+updated 2/6/24
}


\subsubsection{Peaceable Armies of Queens}

This puzzle is also named `Peaceably Coexisting Armies of Queens'~\shortcite{Smith2004:models}.
The problem asks how to place two equal-sized armies of queens on a chessboard so that the white queens do not attack the black queens, and vice versa.
On a standard 8 by 8 board, there are 71 non-isomorphic solutions and their number grows quickly with board size.
Some early results and discussion are due to Stephen Ainley~\citeyear[Problem C5]{Ainley1977:mathematical}. A sophisticated model that makes use of the \textit{regular} constraint was used in the MiniZinc Challenge 2022. 

We use a very simple model, based on the \textit{Basic Model} of \citeA{Smith2004:models}, that takes a single parameter \(n\) for the board size (defining an \(n\times n\) board). For each square on the board we have a variable with domain \(\{0,1,2\}\) with the values indicating no queen, white, and black respectively.
As in the third model of Smith et al., we also use one additional variable \textit{armySize} to indicate the number of queens in each army, and this is maximised.
There are two \code{sum} constraints stating that there are \code{armySize} occurrences of values 1 and 2 respectively. The vast majority of the constraints are to prevent an attack between a pair of queens from opposing armies. Supposing the board is named \code{b}, and squares \((i,j)\) and \((k,l)\) share a row, column, or diagonal, then the \textit{attack} constraint \code{b[i,j]+b[k,l] != 3} is posted. 

We experimented with instances where \(n\in\{4\ldots 11\}\). 
The two army size constraints are identified (together) by the Identical Scopes heuristic. For \(n\geq 5\) the candidate has more than 20 variables and is discarded, otherwise tabulation fails the progress check after 1,000 nodes. Each attack constraint is identified by the Weak Propagation heuristic and all are tabulated (creating binary table constraints). The attack constraints are identical apart from the variable names, so after the first is tabulated the rest are retrieved from the cache. Without tabulation a new integer variable is introduced for each sum \code{b[i,j]+b[k,l]} in an attack constraint. 
Note that the MiniZinc Challenge 2022 model has the same viewpoint and the non-attack requirements are implemented with \textit{regular} constraints which would usually propagate strongly, suggesting that tabulation of the attack constraints may be useful. 

\comment{
Checked+updated 1/6/24
}


\subsubsection{Strong External Difference Families}

A Strong External Difference Family (SEDF) is an object defined on a group, with applications in communications and cryptography \cite{Paterson2016:combinatorial}. For the purposes of this paper, a group is a set \(G\) with an associative and invertible binary operation \(\times\). Also, \(G\) must contain an identity element \(e\), which means that \(e \times g = g\) for every $g \in G$. The model of SEDF described below (with additional symmetry breaking constraints) was used to find a number of previously undiscovered SEDFs, including the first in non-Abelian groups \shortcite{Huczynska2021}.

Given a finite group \(G\) on a set of size \(n\), an \((n,m,k,\lambda)\) SEDF is a list \(A_1,\dots,A_m\) of disjoint subsets of size \(k\) of \(G\) such that, for all \(1 \leq i \leq m\), the multi-set \(M_i = \{xy^{-1} \mid x \in A_i, y \in A_j, i \neq j\}\) contains \(\lambda\) occurrences of each non-identity element of \(G\). 

The parameters of the SEDF problem are \((n,m,k,\lambda)\), the group \(G\) given as a multiplication table \code{tab} (which is an \(n \times n\) matrix of integers), and \code{inv}, a one-dimensional table which maps each group element to its inverse. The SEDF is represented as an \(m \times k\) matrix \code{sedf}. The entire \code{sedf} matrix is contained in a single \code{allDiff} constraint. It has row symmetry (as the sets are not ordered) \shortcite{Flener2002:breaking} and also each row has symmetry (as each row represents a set): any two rows may be exchanged in a solution while preserving solutionhood; also any two elements within a row may be exchanged while preserving solutionhood. The variables within each row and the first column are ordered with \code{<} constraints to partially break the two kinds of symmetry. 

To ensure each multi-set \(M_i\) has \(\lambda\) occurrences of each non-identity element, we use the \code{gcc} on the comprehension below. Value 1 has cardinality 0, and all other values have cardinality \(\lambda\). Each \code{gcc} contains all variables in \code{sedf}. Note that both \code{sedf[i,p]} and \code{sedf[j,q]} are individual variables, and that the inner expression is equivalent to \code{tab[X,inv[Y]]} for integer variables \code{X} and \code{Y}. 

\begin{lstlisting}[language=eprime]
[ tab[sedf[i,p], inv[sedf[j,q]]] | 
        p:int(1..k), q:int(1..k), j:int(1..m), j!=i ]
\end{lstlisting}

We use a set of 12 instances where \(19\leq n \leq 29\), \(3\leq m \leq 8\), and \(2\leq k \leq 5\). The instances originate from research into SEDFs~\cite{Huczynska2021} and are provided in the experimental repository. 

The Identical Scopes heuristic identifies the \code{allDiff} and all \code{gcc} constraints together. Duplicate Variables, Large AST, and Weak Propagation heuristics identify each \code{gcc} individually. In each case tabulation fails the progress check at 1,000 nodes. 
Identical Scopes (Integer) identifies cases where \code{tab[X,inv[Y]]} within a \code{gcc} overlaps with \code{X<Y} or \code{Y<X} ordering the first column (where \code{X} and \code{Y} are variables in the \code{sedf} matrix); Large AST (Integer) and Weak Propagation (Integer) identify the rest of the \code{tab[X,inv[Y]]} expressions and all are successfully tabulated for all instances (creating \code{table} constraints of arity 3). 

\comment{
Checked+updated 1/6/24
19_1_C19_3_3_1.csv.param
IS(Integer) when there is a less-than overlapping. 
LAST, WP(Integer) otherwise, for each thing in the GCC. 

29_1_C29_8_2_1.csv.param same. 
}


\subsection{Summary of Results}\label{sub:feas-results-summary}

We have shown the utility of all four TabID heuristics when applied to top-level constraints, and in most cases their nested and integer versions as well. \Cref{tab:heur-per-problem} summarises the set of heuristics applied to each problem class.  The table indicates when a heuristic fires for at least one instance of a problem class, \textit{and} at least one of the subproblems identified by the heuristic is tabulated.  
All heuristics except Duplicate Variables (Nested) and Duplicate Variables (Integer) meet these criteria for at least one problem class. The top-level heuristics are most important and fire on all but one of the problems, while the other heuristics are also contributing and are essential for SEDF. Nested and Integer heuristics tend to become relevant when top-level constraints are beyond the reach of tabulation, for example in SEDF and the largest instances of Knight's Tour.

\begin{table}[t!]
\begin{tabular}{@{}lcccccccccccc}
\toprule
Problem & \multicolumn{4}{c}{Top-level}    & \multicolumn{4}{c}{Nested}   & \multicolumn{4}{c}{Integer} \\ 
\cmidrule(lr){2-5}\cmidrule(lr){6-9}\cmidrule(lr){10-13}
           & IS    & DV    & LAST  & WP    & IS    & DV   & LAST  & WP    & IS    & DV    & LAST  & WP    \\
\midrule
BlackHole  &       &       &       & \tick & \tick \\
BPMP       &       & \tick & \tick & \tick & \tick &      & \tick & \tick &       &       &       & \tick  \\
Handball   &       & \tick & \tick & \tick & \tick &      &       &       &       &       &       & \tick  \\
JPEncod.     & \tick & \tick & \tick & \\
MDSL       & \tick \\
SportsSchC &       &       &       & \tick \\
\midrule
Accordion  & \tick & \tick & \tick & \tick &      &       &       &       &      &       &       & \tick \\
Coprime    & \tick &       &       &       &      &       & \tick & \tick \\
KillerSu.  & \tick &       &       & \tick \\
KnTourSeq  &       & \tick & \tick & \tick &      &       &       &       &       &       &       & \tick \\
KnTourSucc &       & \tick & \tick & \tick &      &       &       &       &       &       &       & \tick \\
NLinkedSeq &       & \tick & \tick & \tick \\
NLinkedOpt &       & \tick &       & \tick \\
PAQueens   &       &       &       & \tick \\
SEDF       &       &       &       &       &      &       &       &       & \tick &       & \tick & \tick \\
\bottomrule
\end{tabular}
\caption{Heuristics that trigger for each problem class: Identical Scopes (IS); Duplicate Variables (DV); Large AST (LAST); and Weak Propagation (WP). A tick indicates that the heuristic identifies at least one subproblem for at least one instance of the problem class \textit{and} the subproblem is tabulated. 
}\label{tab:heur-per-problem}
\end{table}

For 9 of the 15 models, the generated tables are predominantly binary. Examples include games (e.g.\ Accordion Patience), maths problems such as Coprime Sets, and puzzles (e.g.\ Peaceable Armies of Queens). There are four examples of tabulation (predominantly) producing ternary table constraints: Handball Tournament Scheduling; Optimal N-Linked Sequences; SEDF; and Sports Scheduling Completion. BPMP has two sets of expressions that are tabulated: the channelling constraints of arity 4, and division and modulo expressions where an arity 2 table is produced. Killer Sudoku has 2, 3, 4, and 5 arity tables matching the sizes of the clues in the puzzle. Finally, with Still Life tabulation produces tables of arity 9, albeit on Boolean variables. The \textit{nodeLimit} parameter is set to a conservative value (100,000), minimising the time cost of tabulation but also limiting the arity of generated tables. However the preponderance of arity 2 and 3 tables comes from the structure of the models rather than the effect of \textit{nodeLimit}.


\section{Experimental Evaluation of TabID}\label{sec:experimental-eval}

Having shown that TabID identifies and tabulates promising subproblems of each of the benchmark problems, we now evaluate whether TabID actually speeds up solving of these problems. We test this hypothesis with two conventional CP solvers (Minion and Gecode, taking advantage of extensive research into table propagators) and two CP solvers with conflict learning (OR-Tools and Chuffed). We also test the hypothesis in a different setting, where problem instances are encoded into SAT and solved with a recent CDCL SAT solver (Kissat). In this case we experiment with two different encodings of the table constraints. We first give experimental details and describe the statistical method, and then look at each of the problems of \Cref{sec:eval-part1} in turn. 
TabID is configured in the same way as in \Cref{sec:eval-part1}: \textit{nodeLimit} is set to 100,000; and the SAT size limit is \textit{not} applied except where stated otherwise. 

\subsection{Experimental Details}\label{sec:experiment-details}

We evaluate tabulation with five solvers in six configurations, as follows: 

\begin{description}
\item[Minion] Version 1.9.2 of Minion~\cite{Gent2006:minion}, with ascending value and static variable orderings. The Trie propagator is used for all table constraints~\fullcite{Gent2007:data}. 
\item[Gecode] Commit \texttt{d1bd874} on the  \texttt{release/6.3.0} branch of Gecode~\cite{gecode-website-mod} with the Compact Table propagator~\cite{ingmar2018making}. The search order is the same as for Minion.
\item[Chuffed] Version 0.13.0 of Chuffed~\shortcite{Chu:chuffed} with free search enabled (i.e.\ the solver is free to use dynamic search ordering heuristics). Chuffed encodes all table constraints into SAT internally: it uses the support encoding for binary table constraints \cite{Gent2002:arc} and the Bacchus encoding otherwise~\cite{Bacchus2007:GAC}. 
\item[OR-Tools] Version 9.7.2996 of OR-Tools~\cite{ortools} CP-SAT solver with free search enabled. The support encoding is used for binary table constraints \cite{Gent2002:arc}. Each non-binary table constraint is compressed to a set of short tuples or c-tuples \cite{Katsirelos2007:compression} which is then encoded using basic Boolean constraints similarly to \shortciteA{shortsupportenc-cp16}.
\item[Kissat] Version 3.1.1 of Kissat~\cite{kissat} with \savilerow's default SAT encodings for all constraint types unless otherwise stated. 
The support encoding is used for binary table constraints \cite{Gent2002:arc} and the Bacchus encoding for other tables~\cite{Bacchus2007:GAC}. 
\item[Kissat (MDD)] As above, but non-binary tables are compressed into Multi-valued Decision Diagrams (MDDs) which are then encoded with the GenMiniSAT encoding~\shortcite{abio2016cnf-mod}. 
\end{description}

The solvers and configurations were chosen to include: the recent Compact Table propagator, as well as an earlier table propagator; conventional CP (Minion and Gecode), clause learning CP, and SAT solvers; and both static and dynamic search orders. We used \savilerow 1.10.0 for the experiments, extended with the new TabID method. 
Each reported time is the median of five runs on a 134-node cluster, each node having two 48-core AMD EPYC3 processors and 512 GB RAM; jobs were submitted requiring 1 CPU core and 6 GB RAM. A time limit of 1 hour was applied. Reported times are total times (unless otherwise stated) and include the time taken by \savilerow to tailor the instance, the time taken to apply TabID (if activated), and the backend solver time. 
Software, models and parameter files for all experiments are publicly available online \cite{software-release-tab}.

\subsection{Statistical Analysis}\label{sec:stats}
\begin{figure}[tbp]
\begin{center}
\includegraphics[width=\textwidth]{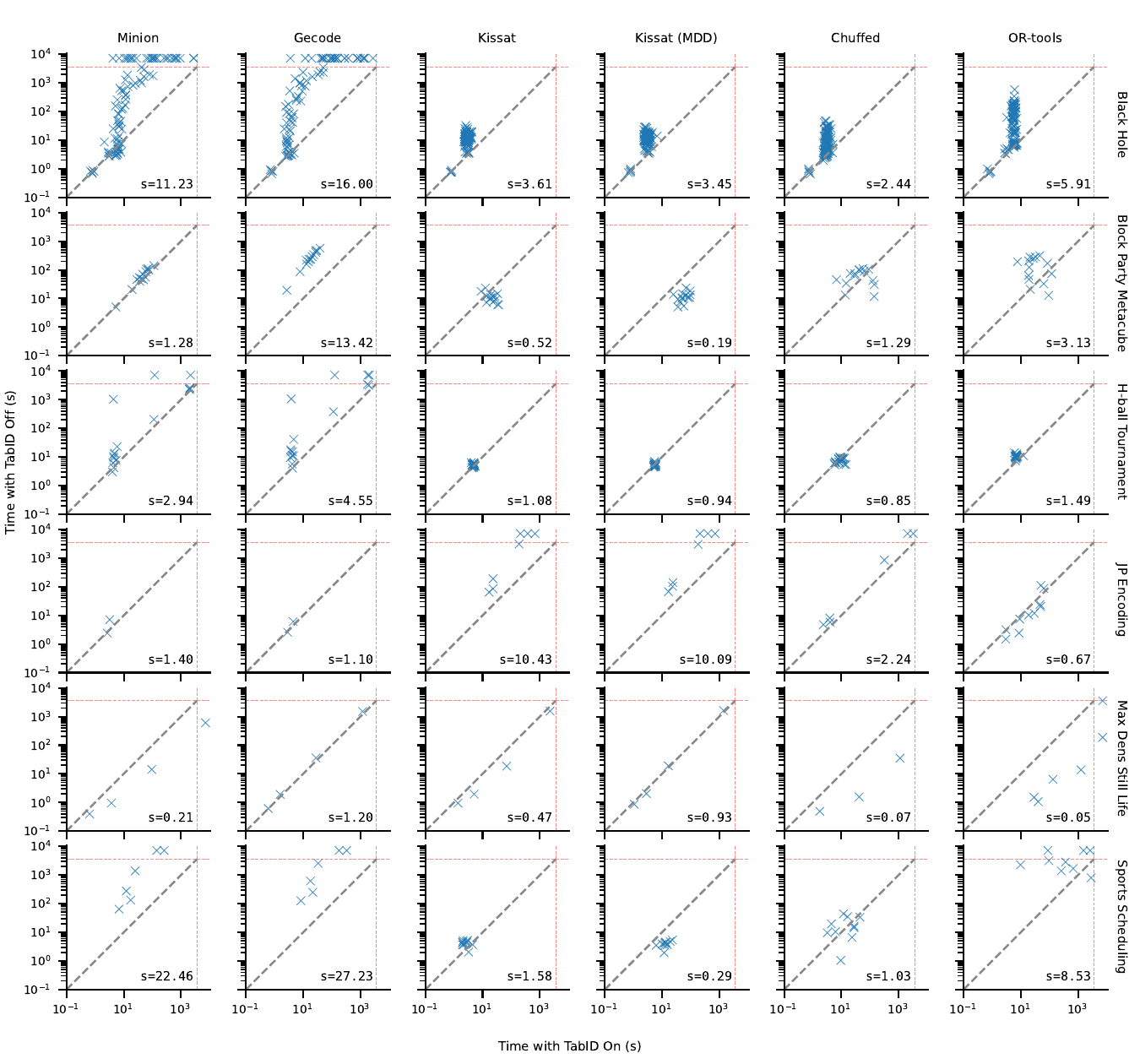}
\end{center}
\caption{\label{fig:grid-dek}\textbf{TabID without SAT Size Limit vs No TabID.} Total runtime (including both \savilerow and the solver) with each of the six solvers for the problem classes: Black Hole, Block Party Metacube Problem, Handball Tournament Scheduling, JP Encoding, Maximum Density Still Life, and Sports Scheduling Completion. 
Time with TabID is shown on the \(x\)-axis and without TabID on the \(y\)-axis. The red dotted lines indicate the time limit of 1 hour; points beyond the line timed out and had the PAR2 penalty applied. Points above the \(x=y\) diagonal were solved faster with TabID than without. The \(s\) value on each plot is the geometric mean of speedup quotients.}
\end{figure}

To compare two configurations A and B of \savilerow (for example, where A does not include TabID and B does), we first take the median of the total time for each instance and each configuration. The median was chosen because it is less affected by outliers than the mean. Instances where \textit{both} configurations timed out are discarded. For the remaining timeouts (i.e. median total time is \(>\)3600 s) we apply a PAR2 penalty, so their median total time is considered to be 7200 s. 
For each instance, we take the quotient of the two medians (\(\frac{A}{B}\)). We take the geometric mean of the set of quotients to obtain \(s\), a single statistic to compare A and B. If \(s>1\) then B is considered to be better than A. The geometric mean is more appropriate than the arithmetic mean in this case~\cite{sr-journal-17}. 

Where \(s\) is close to 1, it may not be clear whether the difference between A and B is statistically significant. As in \citeA{sr-journal-17}, we use the bootstrap method to compute a 95\% confidence interval of \(s\) with 100,000 bootstrap samples. We consider the difference between A and B to be statistically significant when the 95\% confidence interval does not include 1. 

\subsection{Evaluation of Baseline Problems}\label{sec:exp-baseline}

\begin{figure}[t!]
\begin{center}
\includegraphics[width=\textwidth]{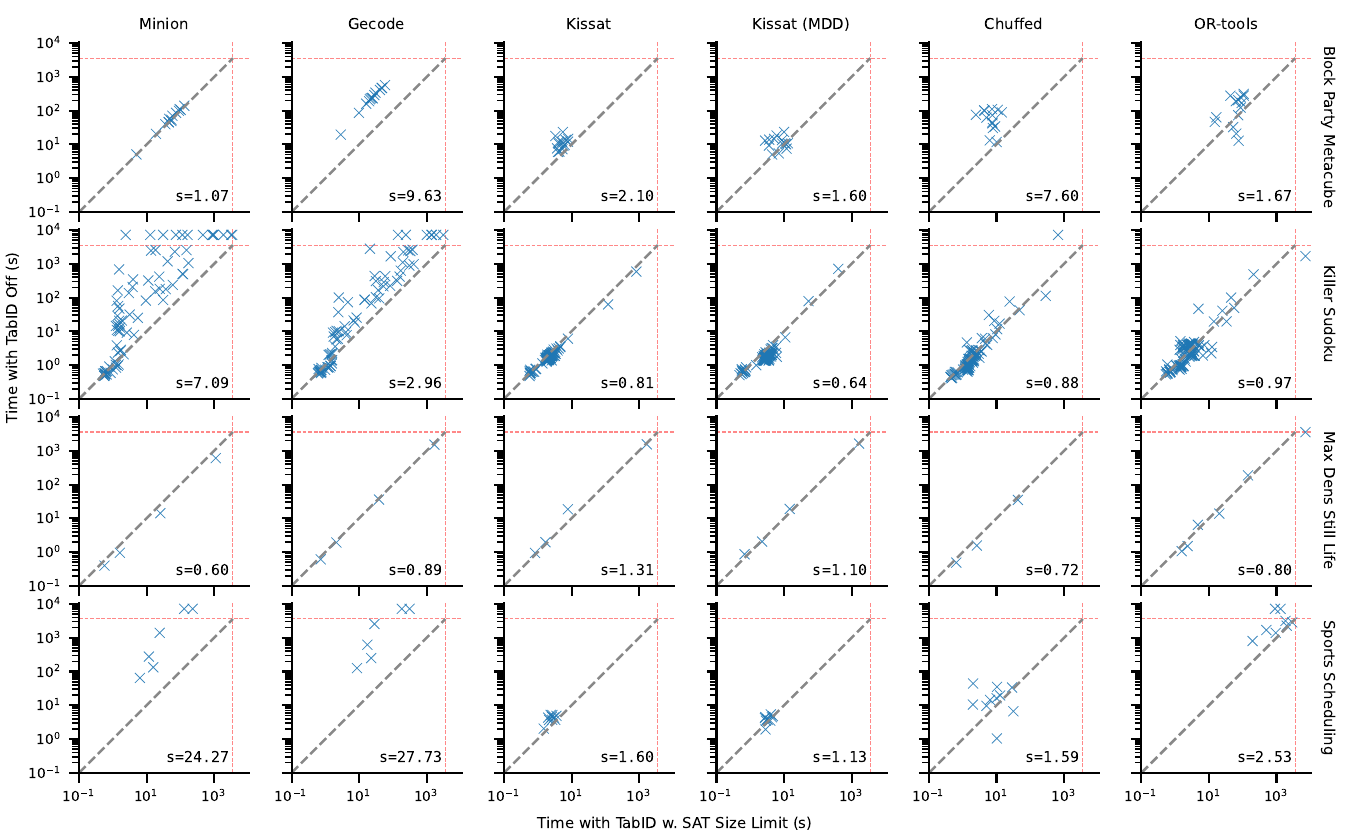}
\end{center}
\caption{\label{fig:grid-results-satlim-wins}\textbf{TabID with SAT Size Limit vs No TabID.} Total runtime (including both \savilerow and the solver) with each of the six solvers for the problem classes:  Block Party Metacube Problem, Killer Sudoku, Maximum Density Still Life and Sports Scheduling.  Time with TabID using the SAT size limit is shown on the \(x\)-axis and without TabID on the \(y\)-axis. Other details are the same as in \Cref{fig:grid-dek}.}
\end{figure}

First we present results for six problem classes where table constraints have been applied in the literature to improve a constraint model. 
Four of these classes were used in~\citeA{Dekker2017:autotabling} to evaluate their auto-tabling method, while the other two are from earlier constraint modelling literature. Results are shown in Figure \ref{fig:grid-dek}. For each problem class we use the set of instances described in \Cref{subsec:eval-part1-baseline}. In this section we give a summary of results. Further analysis can be found in \Cref{appsub:extra-results-baseline}. 

For Black Hole, TabID significantly speeds up all six solvers with the most substantial improvements seen in solvers without clause learning. 
The results are mixed for BPMP, where Minion, Gecode, and OR-Tools show a significant speedup, Chuffed does not reach significance and the SAT solvers are slowed down by TabID. 
For Handball Tournament Scheduling there is a clear benefit for the CP solvers without learning, but for all other solvers the instances are too easy to show a clear difference. 
On the JP Encoding problem, TabID substantially improved the SAT solver's performance but made much less difference with the CP solvers. 
Maximum Density Still Life might seem an obvious candidate, however we found that TabID does not strengthen propagation (node counts for Minion and Gecode are not reduced) and results are mostly negative.  Finally, for Sports Scheduling TabID substantially improved the performance of the CP solvers without learning (as expected) but results with the learning solvers are mixed. 

For problems where there are substantial negative results (\(s<0.5\) for any solver), adding the SAT size limit can help. Results with the SAT size limit enabled are shown in \Cref{fig:grid-results-satlim-wins}. For BPMP, we found that adding the SAT size limit improves performance for Kissat, Kissat (MDD), and Chuffed. For Maximum Density Still Life and Sports Scheduling, switching on the SAT size limit broadly improves the performance of TabID.

\subsection{Evaluation of New Case Studies}\label{sec:exp-newproblems}
 
\begin{figure}[tb]
\begin{center}
\includegraphics[width=\textwidth]{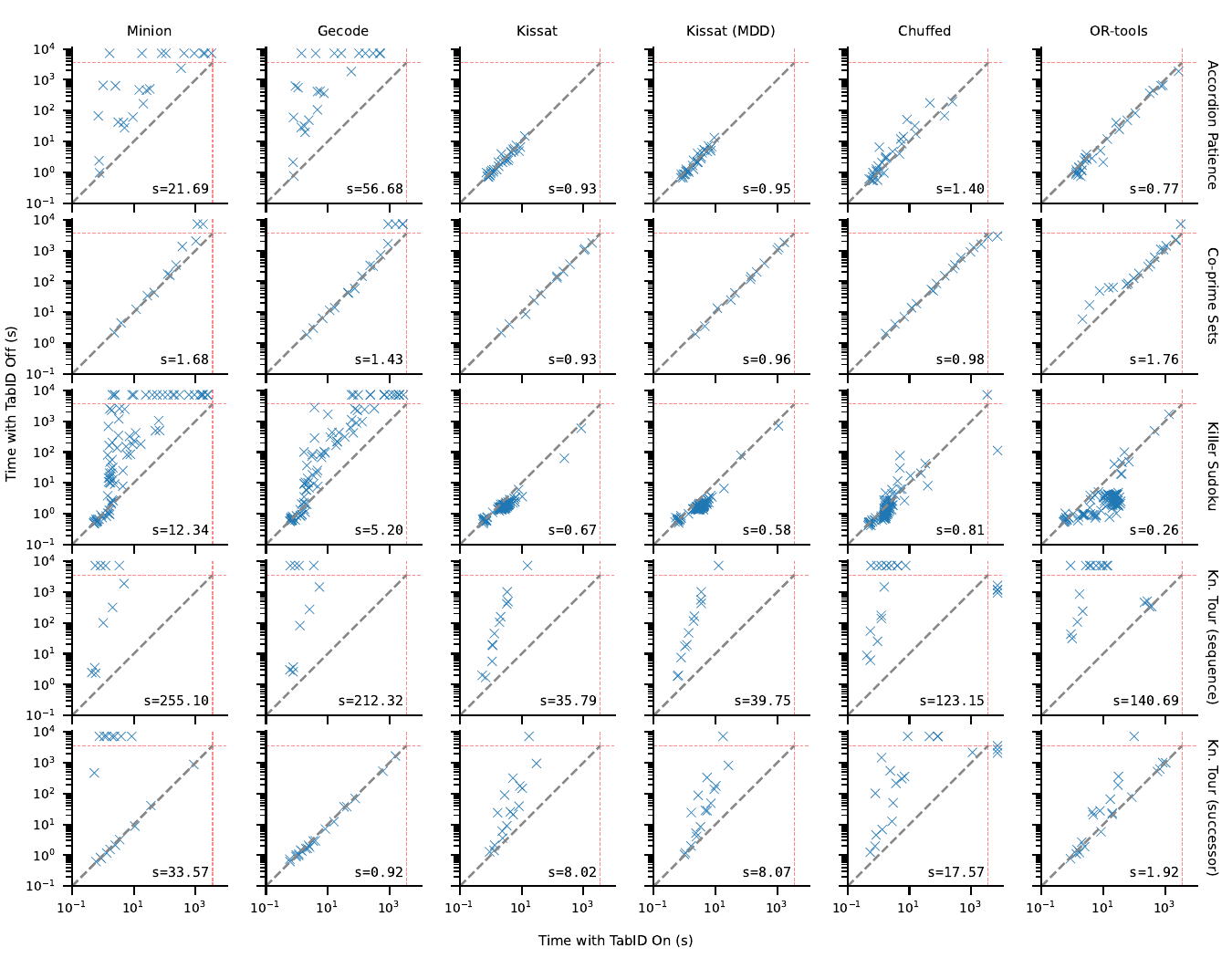}
\end{center}
\caption{\label{fig:grid-results-a}\textbf{TabID without SAT Size Limit vs No TabID.} Total runtime (including both \savilerow and the solver) with each of the six solvers for the problem classes: Accordion Patience, Coprime Sets, Killer Sudoku, and Knight's Tour (with two models). Time with TabID is shown on the \(x\)-axis and without TabID on the \(y\)-axis. Other details are the same as in \Cref{fig:grid-dek}.
}
\end{figure}

\begin{figure}[t]
\begin{center}
\includegraphics[width=\textwidth]{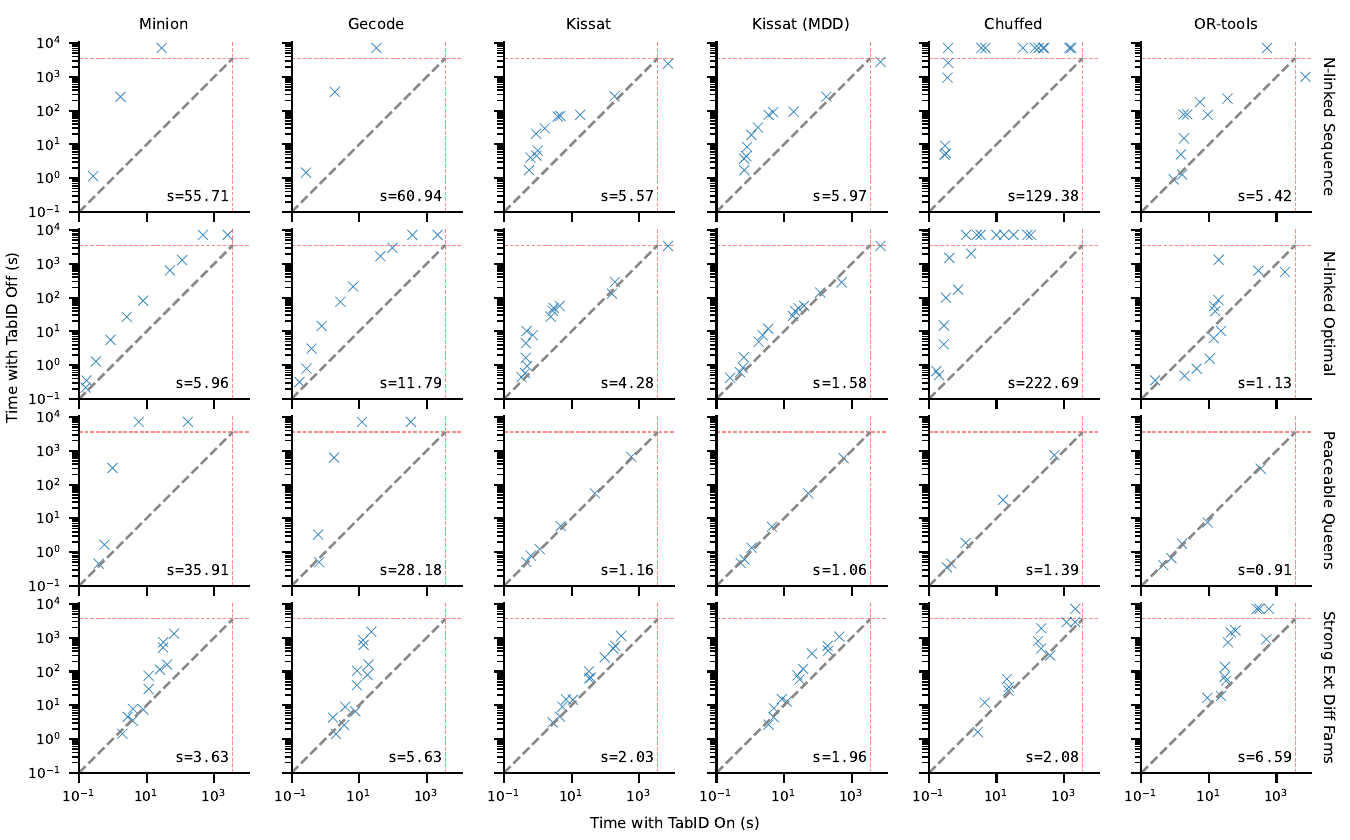}
\end{center}

\caption{\label{fig:grid-results-b}\textbf{TabID without SAT Size Limit vs No TabID.} Total runtime (including both \savilerow and the solver) with each solver for the problem classes: N-Linked Sequence, Optimal N-Linked Sequence, Peaceable Armies of Queens, and Strong External Difference Families.  Time with TabID is shown on the \(x\)-axis and without TabID on the \(y\)-axis. Other details are the same as in \Cref{fig:grid-dek}.}
\end{figure}

In this section we give experimental results for seven new case studies described in \Cref{sec:exp2}. 
Results are shown in Figures \ref{fig:grid-results-a} and \ref{fig:grid-results-b}, and for all problem classes the set of instances is described in \Cref{sec:exp2}. 
We give an overview of results here: further discussion of results can be found in \Cref{appsub:extra-results-newcase}.

For Accordion Patience, TabID dramatically reduces search with the non-learning CP solvers, however with the learning solvers the difference is small or even negative. On the Coprime Sets problem, TabID allows a few additional instances to be solved by the non-learning CP solvers but makes little difference to the other solvers despite TabID changing the model extensively. On the Killer Sudoku problem, TabID strongly improves performance of the non-learning CP solvers but also increases the SAT encoding size quite substantially and slows down all learning solvers on average. Adding the SAT size limit moves all the average speedups closer to 1 for Killer Sudoku (\Cref{fig:grid-results-satlim-wins}). On the \textit{sequence} model of Knight's Tour, TabID causes a very clear speedup for all solvers. When \(n\leq 15\) TabID produces an improved model with much stronger propagation and no auxiliary variables. With the \textit{successor} model the results are mainly positive but complex, and the reasons for that are discussed in \Cref{appsub:extra-results-newcase}. 

TabID speeds up all solvers quite substantially on the N-Linked Sequence problem, however Kissat and OR-Tools mainly benefit for the easier instances suggesting these solvers are able to improve the formulation over time by conflict learning. 
Results for Optimal N-Linked Sequence are similar but with reduced speedups due to the generation of larger tables. 
For Peaceable Armies of Queens, the tabulated model propagates much better leading to big speedups for the non-learning CP solvers, however the learning solvers are almost unchanged. 
Finally, SEDF demonstrates that large speedups are possible (for all solvers) even when no top-level constraints are tabulated.

\subsection{Discussion}\label{sec:exp-results-summary}

With the conventional CP solvers Gecode and Minion, we expected TabID to improve solving time, and it does in the vast majority of cases. Three of the four groups of heuristics (Identical Scopes, Duplicate Variables, Weak Propagation) are based on strength of propagation, i.e.\ identifying sub-problems where propagation is expected to be weak. Assuming the heuristics are accurate, the question then is whether the benefit of obtaining GAC outweighs the cost of generating and propagating table constraints. Gains range from minor efficiency improvements (e.g.\ Block Party Metacube Problem with Minion) to dramatic reductions in search (e.g.\ Black Hole, Killer Sudoku, N-Linked Sequence). 
The other group of heuristics (Large AST and its integer and nested versions) aim to improve efficiency by replacing a large, cumbersome expression with a table. Large AST heuristics are triggered on several problem classes but typically fire together with other heuristics (see \Cref{tab:heur-per-problem}). One example is SEDF, and in this case TabID is worthwhile (particularly for the most difficult instances of SEDF which are sped up by over 40 times). 
We also have a few cases where TabID slows down search: the Knight's Tour \textit{successor} model with Gecode (because Gecode has an efficient GAC propagator for the shift constraint), and Still Life with Minion (where TabID does not strengthen propagation and the table constraints are simply slower to propagate). 

For the solvers with conflict learning (Kissat, Kissat (MDD), Chuffed, OR-Tools) the picture is not as simple. None of them use a native table propagator. Table constraints are encoded into SAT and the size of the encoding affects solver performance. Also, conflict learning may be able to mitigate weak propagation. For some problem classes, TabID is clearly worthwhile: for Black Hole, the solving time becomes approximately constant and the most difficult instances (without TabID) are sped up by approximately 10 times for the SAT solvers and more for Chuffed and OR-Tools; both models of Knight's Tour are sped up by orders of magnitude (except the successor model with OR-Tools); and N-Linked Sequence exhibits large speed-ups for some instances. In each case the generated table constraints are binary and are encoded compactly without any additional SAT variables. SEDF (where the generated table constraints have arity 3) also benefits from TabID to a smaller degree. 

All problem classes where TabID is notably worse with the conflict learning solvers have non-binary constraints (BPMP for some solvers/instances, Sports Scheduling Completion for some solvers/instances, Still Life, and Killer Sudoku). The results suggest that arity, SAT encoding size, or both are important. Adding the SAT size limit to TabID improves average performance on all four of these problem classes, for most of the conflict learning solvers. The exceptions are BPMP and Sports Scheduling with OR-Tools.  \Cref{fig:grid-results-satlim-wins} shows the performance using TabID with the SAT size limit applied for these four problem classes.  Full results with the SAT size limit are given in \Cref{appendix:satsizelimit}.

Finally, the overhead of TabID proved to be acceptable in these experiments. All runtimes reported in this section are total times, but further analysis presented in \Cref{sec:tab-overhead} shows that the time spent applying TabID (as a proportion of total time) is usually small.


\section{Experimental Evaluation of TabID on Other Problem Classes}
\label{sec:eval-other}

In this section we evaluate TabID on a large set of models and problem instances from an existing collection. The purpose is to evaluate the effect of TabID on models where we do not necessarily expect it to improve the model. The key questions are whether TabID represents an unacceptable overhead, and whether the heuristics and limits in TabID accurately identify only those subproblems where tabulation will be useful. 
The \textit{nodeLimit} parameter is set to 100,000 and the SAT size limit (\Cref{sub:satsizelimit}) is applied for all 6 solvers and all problem classes. 

We used all 50 models and 596 instances from \citeA{sr-journal-17}, including models and instances that were used in other experiments in this paper. The models are of a variety of problems including combinatorial designs (e.g.\ BIBD, OPD), puzzles (such as English Peg Solitaire), and industrial design problems (such as SONET). 
All models and instances are available in the experiments repository. 

\begin{figure}[p]
\begin{center}
\includegraphics[width=0.99\textwidth]{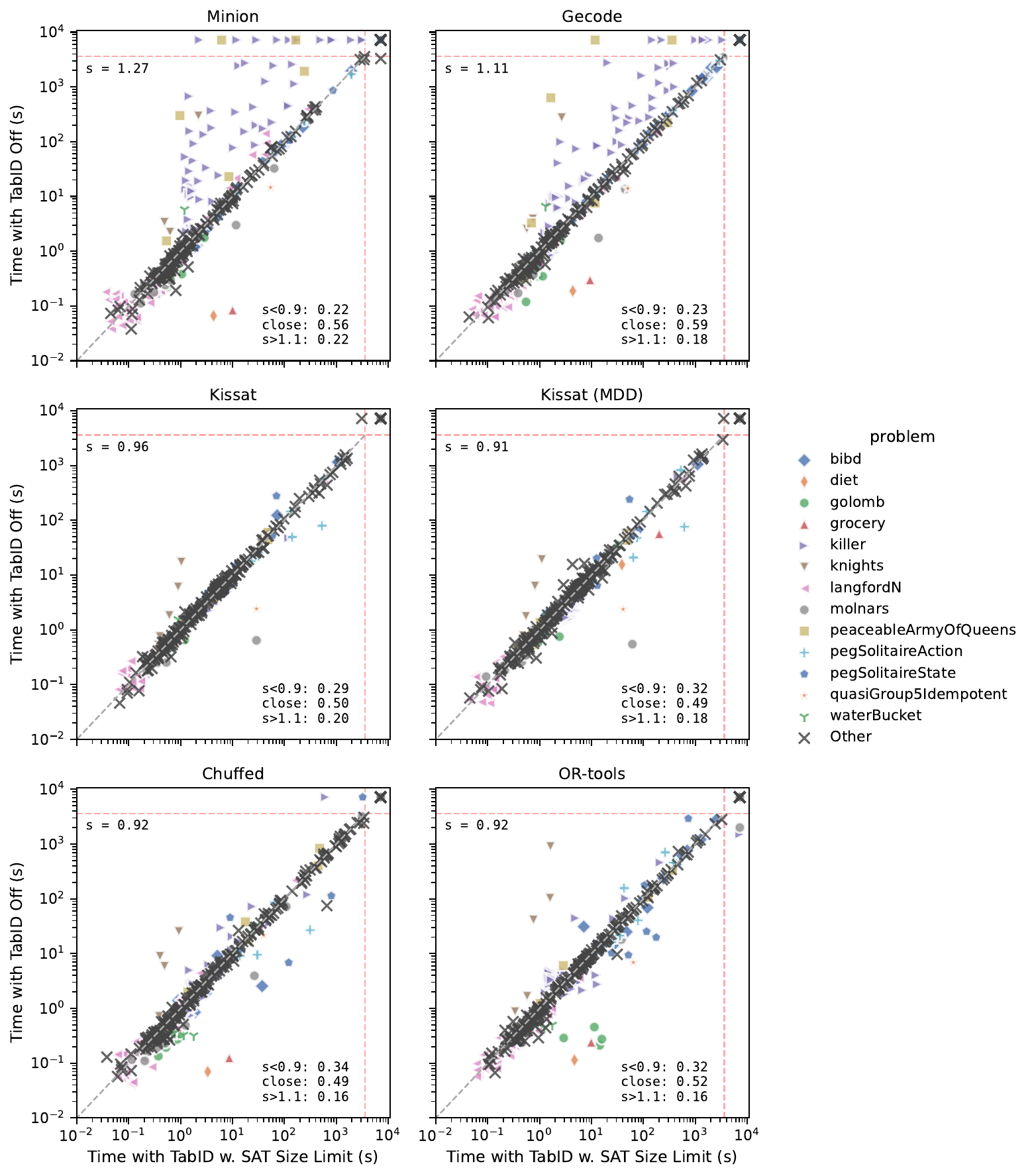}
\end{center}
\caption{\label{fig:others}%
For each of the six solvers, the figure plots the total time (including both \savilerow and the solver) with TabID using the SAT size limit on the \(x\)-axis and without TabID on the \(y\)-axis for a set of 50 models. Other details are the same as in \Cref{fig:grid-dek}. The problem classes with a distinct symbol and colour have at least one point away from the \(x=y\) diagonal on at least one of the plots.}
\end{figure}

Results are plotted in \Cref{fig:others}. In this case we have one plot for each solver, and all 50 models are plotted together. 
TabID is a modest win on average for Minion and Gecode, with Killer Sudoku and Peaceable Armies of Queens showing the largest gains. For the other four solvers, TabID (with the SAT size limit) adds a small overhead on average. Each plot in \Cref{fig:others} is annotated with the proportion of instances where speedup \(s\) of the instance is less than 0.9, where it is close to 1 (`close'), and where it is greater than 1.1. Approximately half the instances lie close to the \(x=y\) line in each case (i.e. \(0.9 \leq s \leq 1.1\)). 

The geometric mean speedup for each solver is close to 1, showing that the overhead of TabID does not deteriorate performance excessively. For Gecode and Minion, a modest overall speedup is observed, with 95\% confidence interval values of \([1.05, 1.18]\) and \([1.17, 1.39]\) respectively.  The learning solvers incur a small overhead, with intervals of  \([0.88, 0.96]\) for Chuffed, \([0.94, 0.99]\) for Kissat, \([0.88, 0.94]\) for Kissat (MDD), and \([0.87, 0.96]\) for OR-Tools.  However, we should not simply turn off TabID whenever the chosen solver has conflict learning. Results presented in \Cref{sec:experimental-eval} show that large improvements in performance are possible with conflict learning solvers.

\section{Scalability of TabID}
\label{sec:scalability}

In this section we investigate how our approach scales as the arity of a generated table increases. The SAT size limit is switched off and \textit{nodeLimit} is set to 100,000 for this experiment. Even though large-arity tables may not be of practical interest in general, there might be cases where they are useful and we want our method to successfully generate such tables. For this purpose, we use the optimal n-linked sequence problem introduced in \Cref{sec:nlinked} and investigate the time and search nodes required to generate tables of various arities and also the effect of them on a CP solver.

Here we experiment with the instance where \(n=12\), and we scale tabulation up by increasing the number of sequence variables in the table constraint. In particular, we use 2 to 6 adjacent variables in the table, thus table arities \(r\) range from 3 to 7, and we maximize the sequence length from 6 upwards. To achieve this we have written a separate model for each value of \(r\), and in these models the divisibility constraint is extended to cover \(r-1\) sequence variables. For example, when \(r=4\) the divisibility constraint (for each \code{i} in \(\{3\ldots n\}\)) is as follows:

\begin{lstlisting}[language=eprime]
!active[i] \/ ( ((seq[i]%seq[i-1] = 0)   \/ (seq[i-1]%seq[i] = 0)) 
            /\ ((seq[i-1]%seq[i-2] = 0) \/ (seq[i-2]%seq[i-1] = 0)))    
\end{lstlisting}

We also post an \code{allDiff} constraint over the same set of \code{seq} variables:

\begin{lstlisting}[language=eprime]
active[i] -> allDiff([seq[i-2],seq[i-1],seq[i]])
\end{lstlisting}

The Identical Scopes heuristic identifies these two constraints together as a candidate. For each value of \(r\), 6 constraints of arity \(r\) are tabulated, and 5 binary constraints are tabulated between adjacent pairs of variables within the first 6 variables in the sequence (where \code{active[i]} is true and the conjunction decomposes into \(r-2\) binary constraints before tabulation).

\begin{table}
    \centering
    \footnotesize
    \begin{tabular}{ccrrrrrrc}\toprule
Sequence & Arity & \multicolumn{2}{c}{Tabulation}  &  \multicolumn{3}{c}{Time (s)} & Solver &  Auto \\
\cmidrule(lr){3-4} \cmidrule(lr){5-7}
variables & \(r\) & nodes & tuples & Tabulation & SR Total & Solver & nodes & \\ \midrule
--- & --- & ---    & ---        & ---   &  0.22  &   0.36 & 174,171 & --- \\ 
2 & 3 & 724        & 285        & 0.01  &  0.24  &   0.03 &    2435 & Yes \\
3 & 4 & 4,999      & 3,697      & 0.04  &  0.32  &   0.04 &     588 & Yes \\
4 & 5 & 60,454     & 51,689     & 0.23  &  0.89  &   0.37 &     330 & No \\
5 & 6 & 835,789    & 762,213    & 0.85  &  1.52  &   6.34 &     203 & No \\ 
6 & 7 & 12,268,984 & 11,396,505 & 13.88 & 21.57  & 129.98 &     143 & No \\ \bottomrule
\end{tabular}
    \caption{Results of tabulating subsequences of lengths 2, 3, 4, 5, and 6 (generating constraints of lengths 3, 4, 5, 6, and 7) of the optimal N-Linked sequence problem with \(n=15\) and solving with Minion. The first line shows no tabulation. The total node count, number of tuples generated, and time are reported for the tabulation process in columns 3-5. \savilerow total time, solver time and solver nodes are reported in columns 6-8. Finally the Auto column indicates whether tabulation would have completed with the work limit and progress checks switched on.}
    \label{tab:scaling}
\end{table}

The results of the experiments are shown in \Cref{tab:scaling}. The first row corresponds to the case when no tabulation is invoked. The `tabulation nodes' in the third column is the number of nodes generated to tabulate all candidates. The arity \(r\) candidate is tabulated once and the other 5 are retrieved from the cache. \textit{Auto} on the last column says whether automatic tabulation could be done with the progress checks and work limit on.  For the \textit{No} rows, TabID would stop at the first progress check at 1,000 nodes. 

The results show that the number of nodes explored 
decreases significantly as we generate larger tables. However, tabulation explores more nodes, and tabulation, \savilerow and solver times increase. While the stronger inference comes with a cost, the incurred runtimes seem reasonable until we reach 762,213 tuples (arity 6). With 11 million tuples (arity 7), we observe a notable cost in tabulation, but still we are able to generate and use the tables. 

In summary, the overhead of generating the tables scales as expected and is not prohibitive until we have millions of tuples (\(r=7\)). The best balance of propagation strength against overhead is found with 285 tuples (\(r=3\)), while \(r=4\) is the largest size allowed with the progress checks and work limit switched on.

\section{Related Work}
\label{sec:related}

In accordance with reformulating a subset of the problem constraints for stronger and/or cheaper constraint propagation, the Globalizer tool \shortcite{Leo2013:globalizing} of MiniZinc helps detect opportunities to use global constraints in constraint models. For instance, it can detect that a set of disequality constraints can be converted to an all-different  constraint. However, the approach is not completely automatic. First, the user provides some instances, then the tool solves them and analyzes the solutions to find properties that match global constraints. The detected global constraints are provided as a suggestion to the user because they are not guaranteed to be correct for all problem instances. 

The tool proposed by \citeA{Dekker2017:autotabling} can convert any predicate (Boolean function) in MiniZinc into a table constraint, but the user must annotate the predicates to be tabulated. In the same vein, the IBM ILOG CPLEX Optimization Studio software supports \verb|strong| annotations to indicate that the solver should find a precomputed table  constraint corresponding to a specified set of variables; the resulting table constraint is then added to the model as an implied constraint~\cite{ibm2017}. The Propia library performed a similar step for an annotated goal in ECLiPSe~\cite{leprovost1992:domain}. In all of these approaches, the user is required to identify  the promising parts of a given model for tabulation.

\citeA{ek2018automatic} proposed to automatically encapsulate sub-problems of MiniZinc models and to identify the strongest candidates for tabulation using a set of heuristics based on numerical scores. One example is minimising \textit{argument modesty}, a measure of the number and domain size of arguments (decision variables and problem class parameters) of the sub-problem. The implementation builds on MiniZinc auto-tabling \cite{Dekker2017:autotabling}. The goal of the system is very similar to TabID, but the approach is quite different: encapsulation and scoring are applied to the problem class model, and the heuristics are entirely different to ours. The system was evaluated with models of Black Hole patience and the Block Party Metacube Problem (BPMP). The preliminary results reported were not positive, however there are many ways in which this work could be developed further. 

With the same goals as tabulation, researchers have studied replacing a set of constraints with other constraints.  De Uña et al.~\citeyear{Una2018:compiling} considered alternative data structures to store the contents of tables. Specifically, they proposed the use of Multivalued Decision Diagrams (MDDs) and Deterministic Decomposable Negation Normal Forms (d-DNNFs), motivated by the fact that creating a table can be costly when the reformulated subproblem has a large number of solutions. The experimental results show that while building compact structures like MDDs or d-DNNFs can be substantially faster than building tables in certain problem classes, it is not clear which structure yields the best total solving time overall. The paper suggests that a more robust technique could choose between the three structures (table, MDD, d-DNNF) depending on an estimate of the size of the solution set of the subproblem. In any case, the presented approach does not automatically identify candidate subproblems. The heuristics we present in this paper could be used to detect automatically opportunities for generating MDD  and d-DNNF  constraints. 

As an alternative, \citeA{Loffler2020:regularization} considered transforming subproblem constraints to a regular constraint. The experimental results demonstrate that the best total solving times are achieved either by a table constraint or by the combination of regular and table constraints.  Again, there is currently no specific algorithm to detect candidate subproblems automatically. The approach either relies on the heuristics presented in our earlier work~\cite{cp-tabulation-2018} or is applied manually. 

Our approach could also generate tables in other compact  representations where ordinary tuples are replaced by compressed tuples~\cite{Katsirelos2007:compression,Regin2011:improving,xia2013optimizing}, short tuples \shortcite{Nightingale2011:exploiting,Nightingale2013:short,Jefferson2013:extending}, or smart tuples \shortcite{Mairy2015:smart}. The latter generalizes classical, compressed and short tuples, can lead to exponentially smaller tables, and can encode compactly many constraints, including a dozen  well-known global constraints. Under a very reasonable assumption about the acyclicity of smart tuples, a polynomial time GAC algorithm was introduced and shown to be effective in practice \cite{Mairy2015:smart}. In addition, \shortciteA{Lecharlier2017:automatic} proposed  automatically synthesizing smart table constraints from (ordinary) table constraints. While the theoretical worst-case time complexity of the algorithm is quadratic in the size of the input table, it was shown to have quasi linear execution time on the considered benchmarks. Segmented tuples \shortcite{Audemard2020:segmented} also generalise compressed tuples by allowing tuples to contain sub-tables over a restricted scope. They demonstrate good performance of segmented tables compared to MDD and regular formulations of a crossword design problem. 

Propagation of multiple overlapping table constraints can be stronger than GAC on each constraint individually. For example, \citeA{Schneider2018:extending} proposed a filtering algorithm based on Compact Table for full PWC, a level of consistency that is significantly stronger than GAC. 
More recently \citeA{wangBipEnc2020} presented a decomposition to binary constraints which achieves full PWC in some cases. 

Finally, advances in propagating or encoding table constraints could improve performance of TabID. \citeA{wangAdHoc2023} survey formalisms for ad-hoc constraints (compressed tables, MDDs, among others) focusing on expressive power. SAT encodings continue to improve, e.g.\ \citeA{wangBinConTrees2022} present some faster alternatives to MDD encodings.

\section{Conclusions and Future Work}
\label{sec:conc}

In this paper we have demonstrated that a small set of heuristics can successfully identify promising sub-problems in a constraint model for tabulation, and that these opportunities can be effectively exploited. The entire process is automated in TabID, situated in the constraint modelling system \savilerow. Our heuristics identify the same tabulation opportunities as recent work by \citeA{Dekker2017:autotabling} where manual annotations of a constraint model are used. In addition we have presented nine other case studies demonstrating the efficacy of our heuristics and automated tabulation. We evaluated the method with SAT, learning CP, and conventional CP solvers, on a wide variety of models. The results vary considerably between problem classes, and between solvers on the same problem class. In some cases, the method produces orders of magnitude improvements in solving time and this is achieved completely automatically. In general we have observed more gains with CP solvers that have native, efficient table propagators. When using a SAT solver, the arity of the generated tables and the size of the encoding are important to solver performance. 

We have also evaluated the method on a set of 50 models (from a pre-existing collection) where for most models we expected there to be no opportunities for useful tabulation. In the large majority of cases we found that TabID makes very little difference to total time, the best possible outcome for TabID when there are no opportunities for useful tabulation. In this experiment we applied a SAT encoding size limit in addition to the work limit that is always part of TabID. 
In the final experiment we investigated how tabulation scales as the arity of a constraint is increased, showing (for the Optimal N-Linked Sequence problem) that table generation can scale far beyond the table size (i.e.\ number of tuples) that has the best trade-off between propagation strength and overhead. 
Currently, a potential user of TabID would be well advised to experiment with a few instances of their problem class to determine whether TabID is useful.

There are two main avenues of future work. Firstly, there is opportunity to refine and extend the collection of heuristics in TabID. Machine learning could be employed to help decide which expressions are promising candidates, as well as to tune the parameters of the tabulation procedure. It has already been applied to predict whether a problem class is generally amenable to tabulation~\shortcite{Cena2023:learning}, as well as to predict whether a \texttt{regular} or a \texttt{table} constraint works faster in a CSP with a single constraint~\shortcite{Loffler2021:decision}. Secondly, it would be interesting to investigate generating compressed table representations such as MDDs (with a work limit and progress checks adapted to the representation). For some compressed representations (including MDDs) there already exist algorithms to generate them (surveyed in \Cref{sec:related}). Integrating compressed representations promises to improve the scalability of the method in some cases, allowing more of the identified candidates to be tabulated.

\acks{We thank EPSRC for grants numbered 
EP/P015638/1\comment{pipeline}, EP/P026842/1\comment{graphs}, EP/R513386/1\comment{DTP FVUO}, EP/W001977/1\comment{sloops}, and  EP/V027182/1\comment{keeplearning} which have supported the authors at various times while this research was undertaken. 
Dr Jefferson held a Royal Society University Research Fellowship.
This project was undertaken on the Viking Cluster, which is a high performance compute facility provided by the University of York. We are grateful for computational support from the University of York High Performance Computing service, Viking and the Research Computing team.
}

\appendix

\section{Identifying Promising Nested Expressions}\label{appn:heur-nested}

\subsection{Identifying Promising Nested Boolean Expressions}\label{sub:nested-bool}

In this section we adapt the four heuristics to identify candidates for tabulation from the set of nested Boolean expressions. The heuristics are adapted as follows (in the order they are applied). 

\begin{description}
\item[Identical Scopes (Nested)] identifies a Boolean expression \(c_1\) that has the same scope as a top-level constraint \(c_2\), where \(c_2\) does not contain \(c_1\). 
\item[Duplicate Variables (Nested), and Large AST (Nested)] are unchanged apart from applying to nested Boolean expressions rather than top-level constraints. 
\item[Weak Propagation (Nested)] identifies a Boolean expression \(c_1\) that is likely to propagate weakly (\Cref{sec:strongprop}), and there exists a top-level constraint \(c_2\) that propagates strongly, with at least one variable in the scope of both \(c_1\) and \(c_2\). 
\end{description}

Once again, subproblems containing more than 20 distinct variables are not considered as candidates for tabulation. 
All four heuristics except Identical Scopes (Nested) identify a single Boolean expression to be tabulated. Identical Scopes (Nested) is somewhat different: it identifies a single Boolean expression (named \(c_1\)) to be replaced, but the table constraint is generated from the conjunction of \(c_1\) with all top-level constraints that have the same scope as \(c_1\). 
For example, suppose that \(x\neq y\) is a nested Boolean expression \(c_1\), and \(x\leq y\) is a top-level constraint \(c_2\) (the only top-level constraint with scope \(\{x,y\}\)). \(x\neq y\) would trigger the Identical Scopes (Nested) heuristic. A table would be generated for \((x\neq y) \wedge (x\leq y)\) and \(c_1\) would be replaced with the new table constraint, while \(c_2\) would remain unchanged. 

The nested Boolean expression heuristics are applied after the top-level constraint heuristics. They are applied to the AST in a top-down order: all four heuristics are applied to an AST node \(n\) before any of the descendants of \(n\). The rationale is to tabulate the largest possible Boolean expression to get the most benefit (within the limits described in \Cref{sub:work-limits}). If an expression is identified by one of the heuristics but tabulation fails (e.g.\ by exceeding a work limit), then parts of the expression (i.e.\ descendant nodes in the AST) may still be tabulated. This occurs in the Knight's Tour problem for example, described in \Cref{sec:KnightsTour}.

\subsection{Identifying Promising Integer Expressions}\label{sub:nested-int}

Finally we adapt the heuristics to apply to integer expressions. The integer expression heuristics are applied last. The AST is traversed in the same order as for Boolean nested expressions (i.e.\ parents before children). Tabulating an integer expression has the additional step of introducing a new auxiliary variable. To avoid the overhead of introducing unnecessary variables, we only consider expressions that would be extracted by general flattening (\Cref{sec:intro-ast}). The model may contain identical integer expressions, and to avoid tabulating multiple identical expressions (and introducing multiple auxiliary variables for them) we use a cache mapping expressions to auxiliary variables. Before applying the heuristics to an expression \(e_1\), we first check the cache, and if another expression identical to \(e_1\) has already been tabulated with auxiliary variable \(a_1\) then \(e_1\) is replaced with \(a_1\). 

Prior to applying the heuristics to expression \(e_1\), a \textit{temporary} auxiliary variable \(a_{\mathit{temp}}\) is created, and the constraint \(c_{\mathit{temp}}\): \(a_{\mathit{temp}} = e_1\) is made (but not attached to the AST). In terms of \(e_1\) and \(c_{\mathit{temp}}\), the heuristics are as follows:

\begin{description}
\item[Identical Scopes (Integer)] identifies an integer expression \(e_1\) that contains more than one variable and has the same scope as a top-level constraint \(c_2\) where \(c_2\) does not contain \(e_1\). 
\item[Weak Propagation (Integer)] identifies an integer expression \(e_1\) where either: (a) \(e_1\) is likely to propagate weakly (as in \Cref{sec:strongprop}), and the top-level constraint \(c_2\) that contains \(e_1\) is likely to propagate strongly when \(e_1\) is temporarily replaced with \(a_{\mathit{temp}}\); or (b) the constraint \(c_{\mathit{temp}}\) would trigger the top-level Weak Propagation heuristic. 
\item[Duplicate Variables (Integer), and Large AST (Integer)] are identical to the heuristics Duplicate Variables and Large AST applied to \(c_{\mathit{temp}}\). 
\end{description}

If \(e_1\) triggers any heuristic except Identical Scopes (Integer), tabulation is attempted on the constraint \(c_{\mathit{temp}}\). If tabulation is successful, then \(a_{\mathit{temp}}\) becomes permanent, \(e_1\) is replaced with \(a_{\mathit{temp}}\), and the new table constraint is attached to the AST (as described in \Cref{sec:mechanics}). 
Identical Scopes (Integer) is similar: the only difference is that tabulation is attempted on the conjunction of \(c_{\mathit{temp}}\) with all top-level constraints that have the same scope as \(e_1\). As with top-level and nested heuristics, subproblems containing more than 20 distinct variables (including \(a_{\mathit{temp}}\)) are not considered as candidates for tabulation. 


\section{Generating Tables}\label{sub:gentabs}

The algorithm for generating a table is implemented entirely within \savilerow and operates directly on an arbitrarily nested Boolean expression, avoiding the need to tailor a candidate expression for an external solver.
In preliminary experiments we also implemented tabulation subject to resource limits in Minion, to take advantage of the propagation mechanism in the solver. However, we found that this was slower than our approach of doing tabulation entirely within \savilerow. In our discussion we therefore focus only on the internal generation approach.

Given a Boolean expression \(e\) to tabulate, we first traverse the AST of \(e\) (in depth-first, left-first order) to collect a list of its variables without duplication. 
A table is generated by depth-first search with a static variable ordering (the order of the variable list) and branching on each value in turn. At each node of search the expression is simplified (\Cref{sec:intro-ast}); if it evaluates to \code{false} then the search backtracks (regardless of whether all variables have been assigned). At each leaf node of search that evaluates to \code{true}, we store the corresponding assignment as a tuple in the table. 
This method adapts to the local structure of the expression, and it only considers the remaining subset of variables occurring in a subtree. This is in contrast to a generate-and-test approach to tabulation which would evaluate the expression for all possible assignments to every variable in the expression. 

For example, consider the knight's move constraint of the Knight's Tour \textit{sequence} model with \(n=4\). \Cref{tab:tabulation-ex} shows part of the depth-first search to tabulate the constraint between \lstinline|tour[7]| and \lstinline|tour[8]|, the first pair of variables with domain \(\{1\ldots 15\}\) (i.e.\ all values except the starting position of 0). Once the first variable is assigned to 1, the expression becomes much shorter and simpler. When both variables are assigned, it evaluates to either \lstinline|true| or \lstinline|false|. The first two tuples to be added to the table are \(\langle 1,7 \rangle\) and \(\langle 1,8 \rangle\). The next would be \(\langle 1,10 \rangle\). Tables are generated with tuples in lexicographic order, and with columns in the order of the variable list.
\begin{table}
    \centering
\begin{tabular}{lll}
\toprule
\code{x}\zap{\code{tour[7]} (\code{x})} & \code{y}\zap{\code{tour[8]} (\code{y})} & Simplified expression \\
\midrule
--- & --- & \lstinline!(|x%4-y%4|=1 /\ |x/4-y/4|=2)!\lstinline! \/ (|x%4-y%4|=2 /\ |x/4-y/4|=1)!\\
1 & ---  & \lstinline!(|1-y%4|=1 /\ y/4=2)! \lstinline!\/ (|1-y%4|=2 /\ y/4=1)! \\
1 & 1 & \lstinline|false| \\
\multicolumn{2}{c}{\(\vdots\)}\\
1 & 6 & \lstinline|false| \\
1 & 7 & \lstinline|true| (tuple \(\langle 1, 7\rangle\) added to table)\\
1 & 8 & \lstinline|true| (tuple \(\langle 1, 8\rangle\) added to table)\\
1 & 9 & \lstinline|false| \\
\multicolumn{2}{c}{\(\vdots\)}\\
\bottomrule
\end{tabular}
    \caption{A fragment of the depth-first search to tabulate the knight's move constraint between variables \code{tour[7]} (denoted by \code{x}) and \code{tour[8]} (denoted by \code{y}) of the Knight's Tour \textit{sequence} model, with \(n=4\) and starting location \((0,0)\).}
    \label{tab:tabulation-ex}
\end{table}


\section{Detailed Experimental Results}

In this appendix we provide more detailed analysis including confidence intervals and statistical significance results as described in \Cref{sec:stats}. 
Some problem classes have 10 or fewer instances, in which case we report the speedup quotient for each instance instead of the 95\% confidence interval. 

\subsection{Baseline Problems}\label{appsub:extra-results-baseline}

\subsubsection{Black Hole}

TabID can speed up all six solvers. The solvers without clause learning show very substantial speedups with TabID and the effect is clearly significant. For example, Gecode has a geometric mean speedup of \(16.00\) with 95\% confidence interval \([10.59, 24.01]\). Minion exhibits a small overhead, and some easy instances are slower with TabID than without, but on average there is a large speedup with 95\% confidence interval of \([7.17, 17.73]\). 
The solvers with learning performed much better on Black Hole (with or without TabID), and the average speedup is significant, though smaller: the 95\% confidence intervals for Kissat, Chuffed and OR-Tools are \([3.24, 4.01]\), \([2.09, 2.87]\), and \([4.55, 7.71]\), respectively. The two SAT solver configurations are in fact identical for this problem class because all table constraints are binary and the support encoding is used in both cases (see \Cref{sec:experiment-details}).

\subsubsection{Block Party Metacube Problem}

The solvers exhibit differing speed-ups, with the largest being Gecode where \(s=13.42\) with 95\% confidence interval of \([11.90,14.76]\). For Minion the speed-up is smaller but significant, with a 95\% confidence interval of \([1.14, 1.42]\). 
We found no significant difference with Chuffed, where the 95\% confidence interval is \([0.68, 2.26]\). In the results presented by~\citeA{Dekker2017:autotabling}, Chuffed performs badly without tabulation, whereas here it is solving all instances. Their non-tabulated model has additional \textit{rotation} variables and others (see \Cref{sec:bpmp-eval1}) which may explain the difference in performance. OR-Tools shows a larger (and statistically significant) improvement with confidence interval \([1.44, 6.46]\). 

TabID substantially reduces performance of the SAT solver with both encodings. For this problem the sizes of the SAT encodings are increased by TabID. Adding the SAT size limit (\Cref{sub:satsizelimit}) to TabID improves performance for Kissat, Kissat (MDD), and Chuffed, resulting in average speed-ups of 2.10, 1.60, and 7.60 respectively as shown in \Cref{fig:grid-results-satlim-wins}.

\subsubsection{Handball Tournament Scheduling}\label{sec:hts-eval2}

For this problem class there is a clear benefit for the conventional CP solvers: the 95\% confidence intervals for Gecode and Minion are \([2.49, 9.66]\) and \([1.56, 6.62]\) respectively. For all other solvers, the instances are too easy to show a clear difference. Total times are around 10 seconds for every instance, both with and without TabID. The 95\% confidence intervals for Kissat, Kissat (MDD), Chuffed, and OR-Tools are \([1.02, 1.16]\), \([0.89, 0.99]\), \([0.74, 0.96]\), and \([1.37, 1.62]\) respectively.

\subsubsection{JP Encoding Problem}

For JP Encoding the results are slightly positive for Gecode and Minion, however they can solve only two of the ten instances. Chuffed solves six instances with TabID and four without, and is somewhat faster with TabID. OR-Tools solves all instances with or without TabID, however TabID slows it down on average. 

\savilerow{}'s default SAT encoding for sums does not scale well when encoding the objective function. Instead we used the GGPW encoding \cite{pbamo-aij} applied to the direct encoding of the variables \cite{sr-manual}. 
Kissat and Kissat (MDD) show the largest speed-ups of around 10 times on average. 
The speedup quotients for each solver on each instance (except double timeouts) are shown in the following table. 

{\footnotesize
    \begin{center}
\begin{tabular}{ll}\toprule
Solver & Sorted speedup quotients for tabulation \\ \midrule
Minion & 0.90, 2.18 \\
Gecode & 0.90, 1.35 \\
Kissat & 3.69, 3.82, 8.22, 10.63, 16.49, 19.18, 34.58 \\
Kissat-MDD & 4.03, 4.41, 5.84, 10.04, 16.60, 18.07, 34.04 \\
Chuffed & 1.45, 2.02, 2.06, 2.22, 2.71, 3.51 \\
OR-Tools & 0.28, 0.40, 0.42, 0.49, 0.52, 0.55, 0.81, 1.06, 1.38, 2.23 \\
\bottomrule
\end{tabular}
\end{center}
}

\subsubsection{Maximum Density Still Life}\label{sec:exp-negativeexample}

Still Life might seem to be an obvious candidate for tabulation, however for Gecode and Minion we found that the node counts are identical for all non-timeout instances: TabID does not improve propagation. Minion is substantially slower with TabID, as are the learning solvers except Kissat (MDD). Gecode and Kissat (MDD) have approximately the same performance, as shown in \Cref{fig:grid-dek} and the following table. 

{\footnotesize
    \begin{center}
    \begin{tabular}{ll}\toprule
Solver & Sorted speedup quotients for tabulation \\ \midrule
Minion & 0.08, 0.15, 0.26, 0.62 \\
Gecode & 1.01, 1.23, 1.27, 1.29 \\
Kissat & 0.27, 0.39, 0.68, 0.69 \\
Kissat-MDD & 0.72, 0.80, 1.14, 1.17 \\
Chuffed & 0.03, 0.04, 0.28 \\
OR-Tools & 0.01, 0.03, 0.03, 0.05, 0.05, 0.49 \\
\bottomrule
\end{tabular}
\end{center}
}

Adding the SAT size limit to TabID improves its performance for all solvers except Gecode, resulting in average speed-ups between 0.60 (Minion) and 1.31 (Kissat) as shown in \Cref{fig:grid-results-satlim-wins}. Recall from \Cref{sec:negative-example} that the Identical Scopes heuristic identifies subproblems with arity 9 (for grid squares away from the inner border), arity 6 (for squares adjacent to the inner border), and arity 4 (for corner squares). The SAT size limit always prevents the arity 9 and 6 candidates being tabulated. It also prevents arity 4 candidates being tabulated when the MDD encoding of table is used. Instead, parts of the constraints are identified by Duplicate Variables (Nested) and they are tabulated with mixed results as shown in \Cref{fig:grid-results-satlim-wins}.

\subsubsection{Sports Scheduling Completion}

TabID proves to be highly beneficial for the conventional CP solvers Gecode and Minion, as expected. However, with Chuffed the picture is mixed and the average speedup is very close to 1. Results with Kissat depend on the choice of encoding for the generated table constraints, with the support encoding performing much better than the MDD encoding (however, instances are too easy for firm conclusions). Finally, OR-Tools is less effective than the other learning solvers in general but TabID helps quite substantially. Plots and geometric mean speedups are shown in \Cref{fig:grid-dek}, and the sorted speedup quotients for each solver on each instance (except double timeouts) are shown in the following table. Adding the SAT size limit to TabID improves Kissat (MDD) performance in particular but makes OR-Tools worse (see \Cref{fig:grid-results-satlim-wins}). 

{\footnotesize
    \begin{center}
\begin{tabular}{ll}\toprule
Solver & Sorted speedup quotients for tabulation \\ \midrule
Minion & 7.64, 9.43, 22.48, 27.90, 51.09, 55.63 \\
Gecode & 11.53, 15.19, 21.79, 34.62, 40.22, 76.70 \\
Kissat & 0.63, 0.83, 1.56, 1.72, 1.73, 1.79, 1.87, 2.18, 2.23, 2.47 \\
Kissat-MDD & 0.16, 0.23, 0.24, 0.25, 0.28, 0.32, 0.33, 0.34, 0.38, 0.56 \\
Chuffed & 0.11, 0.28, 0.52, 0.59, 0.75, 1.70, 2.14, 3.00, 3.77, 4.38 \\
OR-Tools & 0.28, 2.51, 2.72, 4.64, 5.37, 7.85, 33.20, 81.58, 234.85 \\
\bottomrule
\end{tabular}
\end{center}
}

\subsection{New Case Studies}\label{appsub:extra-results-newcase}

\subsubsection{Accordion Patience}

For the two non-learning CP solvers, TabID speeds up solving considerably. Gecode has the largest geometric mean speedup of 56.68, with 95\% confidence interval \([25.27, 126.75]\). In some cases TabID reduces search nodes by more than 1,000 times and this translates into very large speedups. For Chuffed the effect is small, the geometric mean speedup is 1.40 with 95\% confidence interval \([1.14, 1.74]\).  TabID seems to cause a slight slow-down for Kissat and Kissat (MDD) but neither reach significance (with 95\% confidence intervals \([0.87, 1.01]\) and \([0.88, 1.03]\) respectively). OR-Tools is also somewhat slower, with confidence interval \([0.67, 0.88]\). 

A preliminary experiment with a particular instance (named \texttt{cards\_11\_01.param}) and Minion (with a static variable and value ordering) showed that almost all the benefit comes from tabulating Move1 and Move2 constraints (described in \Cref{sec:accordion-eval1}). Without tabulation, Minion takes 644,774 nodes. With just Move1 and Move2 tabulated, Minion takes 35,760 nodes, and with TabID it takes 35,712 nodes (and total time was almost identical). 

\subsubsection{Coprime Sets}

For the non-learning CP solvers, TabID is neutral for the smallest (and easiest) instances but starts to pay off for larger instances. The 95\% confidence intervals are \([1.07, 1.98]\) for Gecode and \([1.22, 2.42]\) for Minion. In both cases some additional instances are solved with TabID. 
Kissat and Chuffed show very little difference overall, which we found surprising given the extensive changes that TabID makes to the model. Finally, OR-Tools benefits for easier instances but there is little difference after approximately 100 s. One additional instance is solved with TabID, and the confidence interval is \([1.36, 2.32]\).

\subsubsection{Killer Sudoku}

For the non-learning CP solvers, TabID makes a substantial difference. For Minion the average speedup is 12.34 with 95\% confidence interval \([7.54, 20.47]\). The effect on Gecode is smaller, with geometric mean speedup of 5.20 and 95\% confidence interval \([3.74, 7.28]\). 
However, for all the learning solvers, the geometric mean speedup is below 1 (e.g.\ with Kissat the average is 0.67 with 95\% confidence interval \([0.63, 0.70]\)). In this case TabID increases SAT encoding size quite substantially. Adding the SAT size limit to TabID prevents some of the identified subproblems being tabulated, and for all solvers it moves the geometric mean speedup closer to 1. OR-Tools benefits in particular, with geometric mean speedup of 0.97. The results with the SAT size limit are shown in \Cref{fig:grid-results-satlim-wins}.

\subsubsection{Knight's Tour Problem}\label{sec:knights-eval2}

For the \textit{sequence} model, TabID causes a very clear speedup for all solvers. In this case, tabulation of instances where \(n\leq 15\) produces a quite different model with no auxiliary variables and much stronger propagation. Kissat shows the least improvement, with a geometric mean speedup of 35.79 and 95\% confidence interval \([14.08, 89.36]\). 

With the \textit{successor} model the results are more complex. For Minion, TabID makes little difference except for the 8 instances described in \Cref{sec:knights-eval1} where shift constraints are tabulated. For these 8 instances there is a speedup of more than 100 times. Gecode is slightly slower with TabID; it uses a GAC propagator by default for shift constraints so does not benefit from TabID in this case. The SAT solvers and Chuffed benefit from TabID, in terms of both time and number of instances solved. For example, Kissat has a geometric mean speedup of 8.02 with confidence interval \([3.78, 18.29]\). Finally, with OR-Tools we found that 7 instances were solved faster with TabID and it made almost no difference to the others. The 7 instances are of various sizes from 8 to 15 and have no obvious common features; only three have the shift constraints that made a large difference with Minion.

\begin{figure}[t!]
\begin{center}
\includegraphics[width=\textwidth]{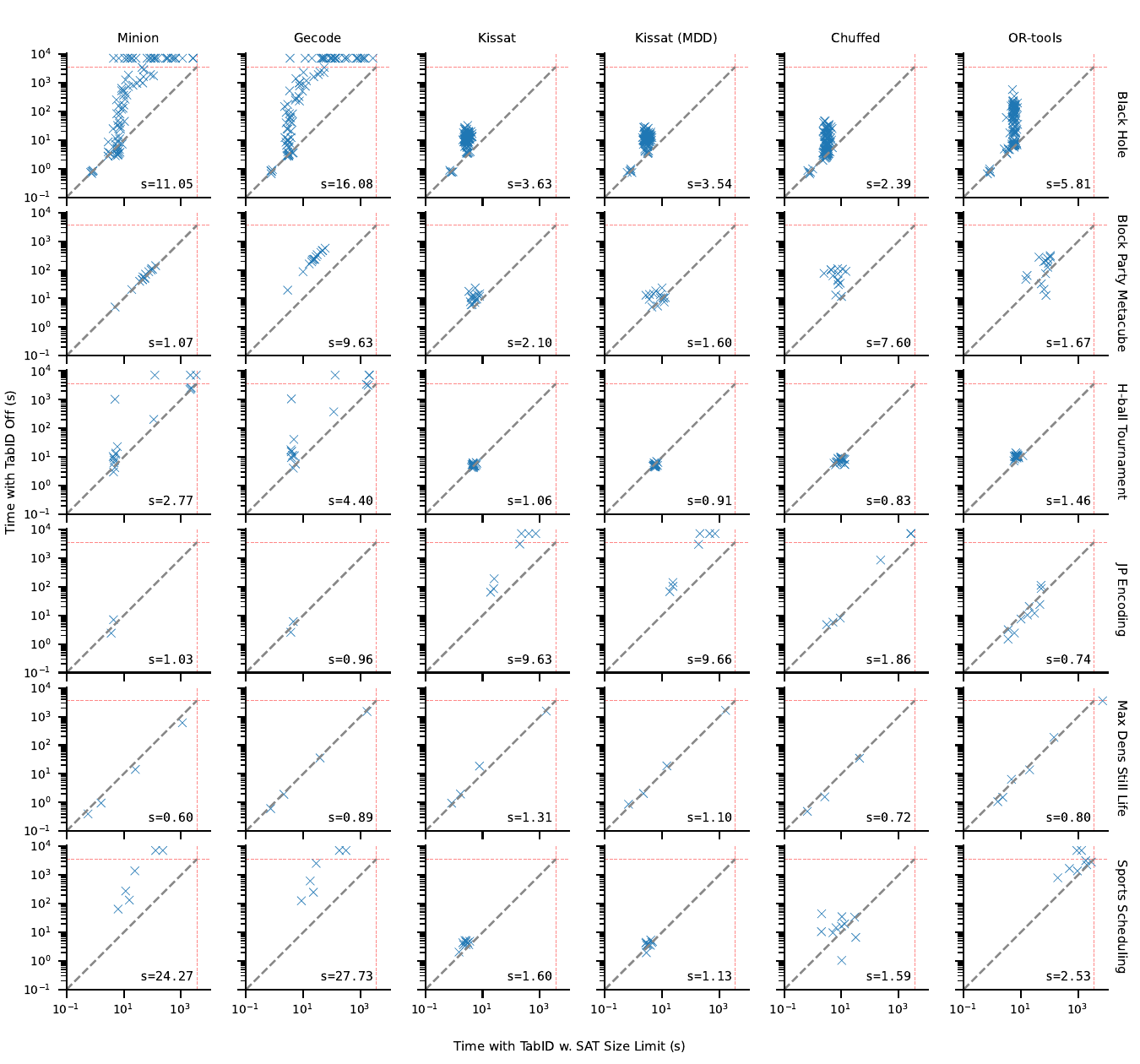}
\end{center}
\caption{\label{fig:grid-dek-tabsatlim}For each of the six problem classes: Black Hole, Block Party Metacube Problem, Handball Tournament Scheduling, JP Encoding, Maximum Density Still Life, and Sports Scheduling, and the six solvers, the figure plots the total time (including \savilerow and the solver) with TabID applying the SAT size limit on the \(x\)-axis and without TabID on the \(y\)-axis. Other details are as in \Cref{fig:grid-dek}.}
\end{figure}

\subsubsection{N-Linked Sequence and Optimal N-Linked Sequence}

For the decision problem, TabID speeds up all solvers quite substantially. For example Chuffed has a geometric mean speedup of 129.38 with 95\% confidence interval \([35.59, 509.32]\). Gecode and Minion struggle to solve many instances with or without TabID. All table constraints are binary so the two SAT configurations are the same (see \Cref{sec:experiment-details}). Kissat and OR-Tools are affected less than the other solvers, but both still reach statistical significance. For example, OR-Tools has confidence interval \([1.85, 13.97]\). For Kissat and OR-Tools, TabID seems to help more on the easier instances, suggesting that these solvers are able to improve the formulation of an instance over time by conflict learning. 

TabID also appears to speed up all solvers for the optimization problem, but for Kissat (MDD) and OR-Tools the gain is small (and for OR-Tools it does not reach significance). Kissat and Kissat (MDD) have 95\% confidence intervals of \([2.18, 8.17]\) and \([1.16, 2.10]\) respectively. Chuffed benefits the most, with 95\% confidence interval \([70.02, 656.66]\). OR-Tools has confidence interval \([0.46, 3.12]\).

The table constraints generated for the optimization problem are of arity 3 and have more tuples for the same value of \(n\). For the decision problem, with \(n\in \{60,70,80\}\), the tables have between 462 and 656 tuples. For the optimisation problem (with \(12\leq n\leq 42\)) the tables have between 202 and 2058 tuples, shifting the trade-off between TabID and the default encodings or propagators. 

\begin{figure}[t!]
\begin{center}
\includegraphics[width=\textwidth]{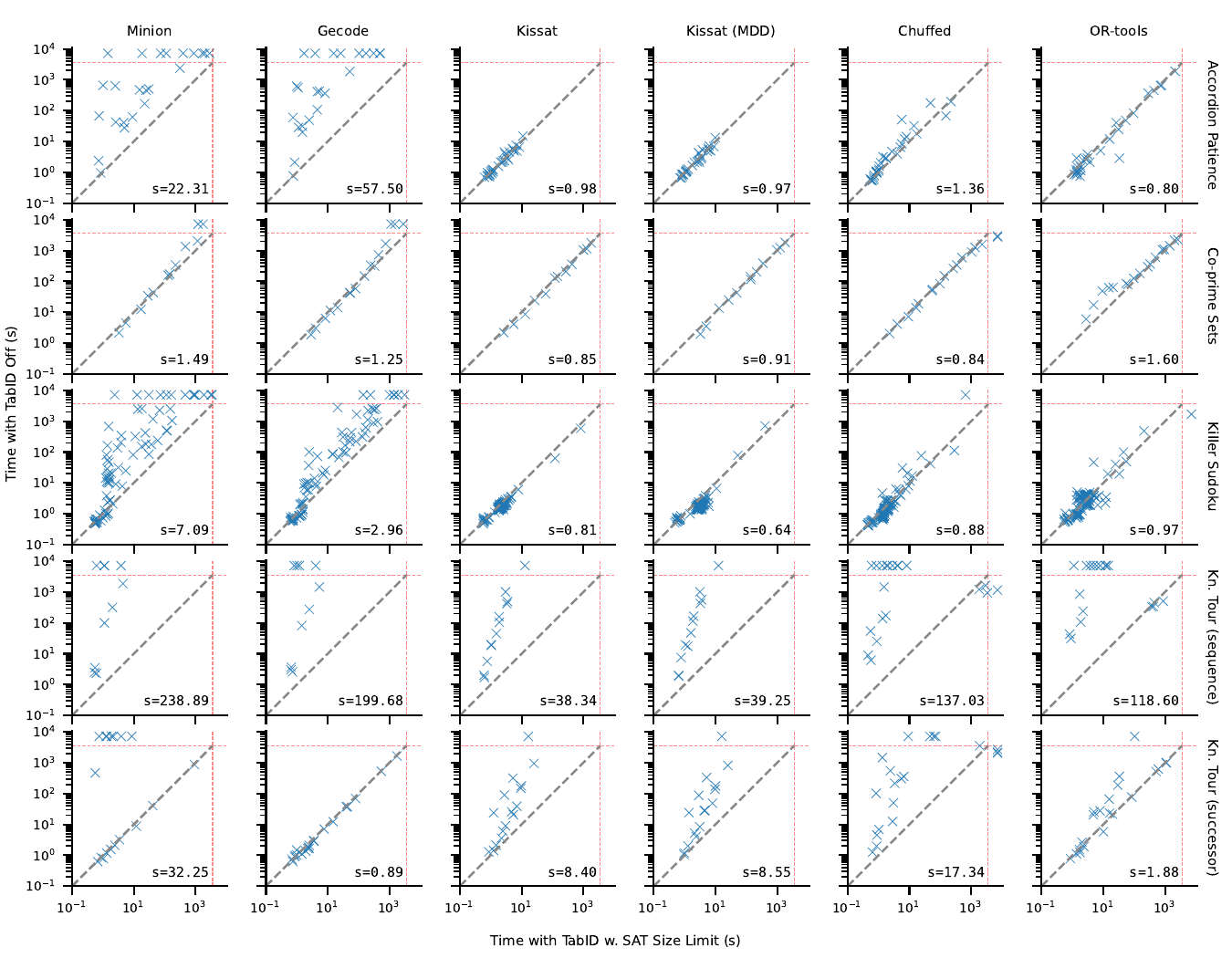}
\end{center}
\caption{\label{fig:grid-results-a-satlim}For each of the four problem classes: Accordion Patience, Coprime Sets, Killer Sudoku, and Knight's Tour (with two models), and the six solvers, the figure plots the total time (including both \savilerow and the solver) with TabID applying the SAT size limit on the \(x\)-axis and without TabID on the \(y\)-axis. Other details are the same as in \Cref{fig:grid-dek}.}
\end{figure}

\subsubsection{Peaceable Armies of Queens}

The tabulated model propagates much better with the conventional CP solvers and they exhibit very large speedups as a result. The conflict learning solver performance is almost unchanged by TabID. Kissat and Kissat (MDD) are identical since all table constraints are binary (see \Cref{sec:experiment-details}). The sorted speedup quotients for each solver on each instance (except double timeouts) are shown in the following table. 

{\footnotesize
    \begin{center}
    \begin{tabular}{ll}\toprule
Solver & Sorted speedup quotients for tabulation \\ \midrule
Minion & 1.19, 2.95, 43.43, 317.85, 1231.42 \\
Gecode & 0.76, 5.38, 21.12, 345.59, 591.67 \\
Kissat & 1.07, 1.11, 1.13, 1.13, 1.24, 1.27 \\
Kissat-MDD & 0.89, 0.91, 1.03, 1.04, 1.23, 1.29 \\
Chuffed & 1.01, 1.05, 1.42, 1.53, 2.23 \\
OR-Tools & 0.83, 0.86, 0.86, 0.92, 1.11 \\
\bottomrule
\end{tabular}
\end{center}
}

\begin{figure}[t!]
\begin{center}
\includegraphics[width=\textwidth]{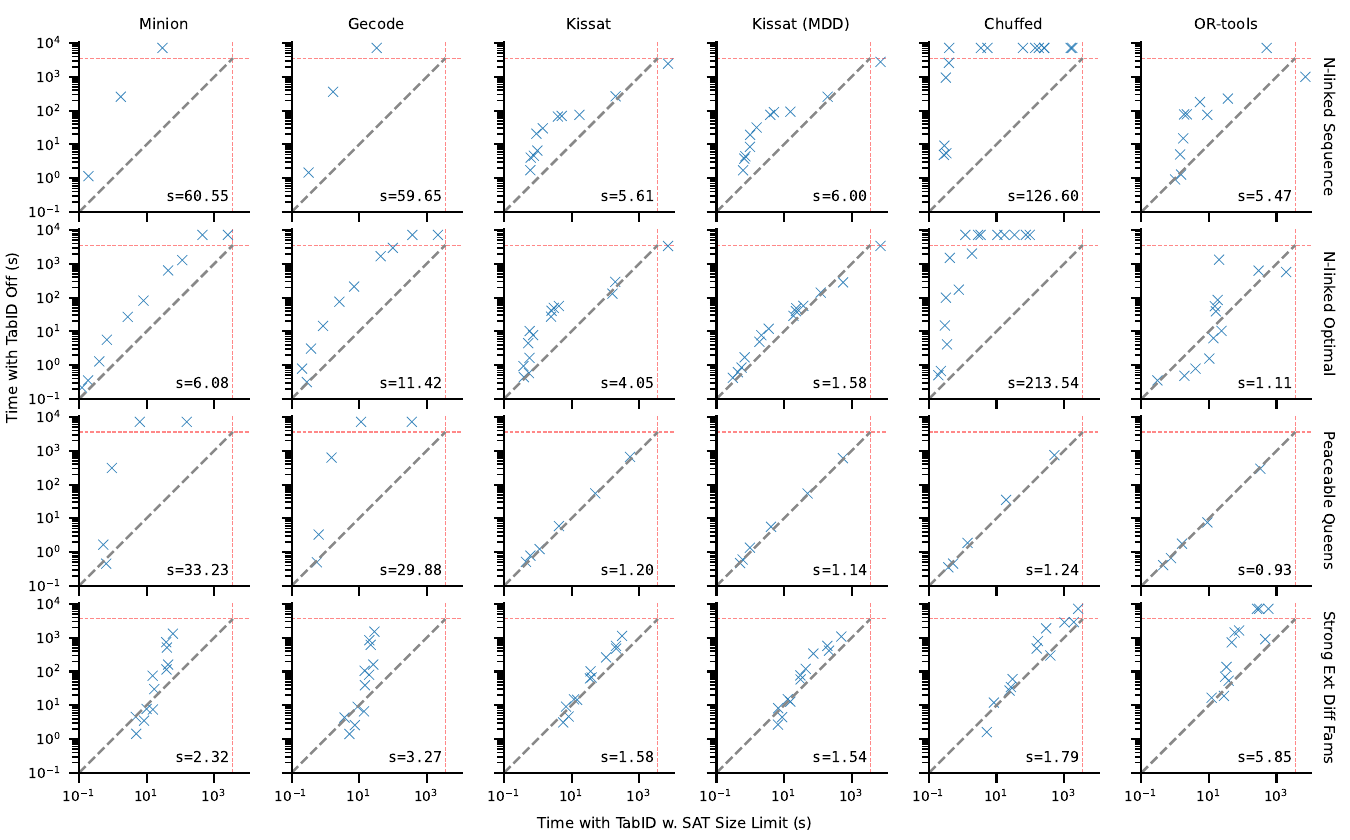}
\end{center}
\caption{\label{fig:grid-results-b-satlim}For each of the four problem classes: (Optimal) N-Linked Sequence, Peaceable Armies of Queens, and Strong External Difference Families (SEDF), and the six solvers, the figure plots the total time (including both \savilerow and the solver) with TabID applying the SAT size limit on the \(x\)-axis and without TabID on the \(y\)-axis. Other details are the same as in \Cref{fig:grid-dek}.}
\end{figure}

\subsubsection{Strong External Difference Families}

The results are positive for all solvers, but particularly for Gecode, Minion, and OR-Tools on larger instances. For example, with Gecode the geometric mean speedup is 5.63 with 95\% confidence interval \([2.29, 14.16]\), and a peak speedup of 65 times. Gecode, Minion, Kissat, and Kissat (MDD) all have no time-outs, with and without TabID. Chuffed solves all instances with TabID, but times out on 1 instance without TabID. OR-Tools is similar but times out on 3 instances without TabID. The 95\% confidence intervals for solvers Chuffed, Kissat, and Kissat (MDD) are \([1.38, 3.16]\), \([1.62, 2.54]\), and \([1.44, 2.64]\) respectively so all reach statistical significance.  
In SEDF all the tabulated expressions are of one kind: they are integer expressions within a \code{gcc} constraint. SEDF demonstrates that large speedups are possible even when no top-level constraints are tabulated. 

\begin{figure}[t!]
\begin{center}
\includegraphics[width=\textwidth]{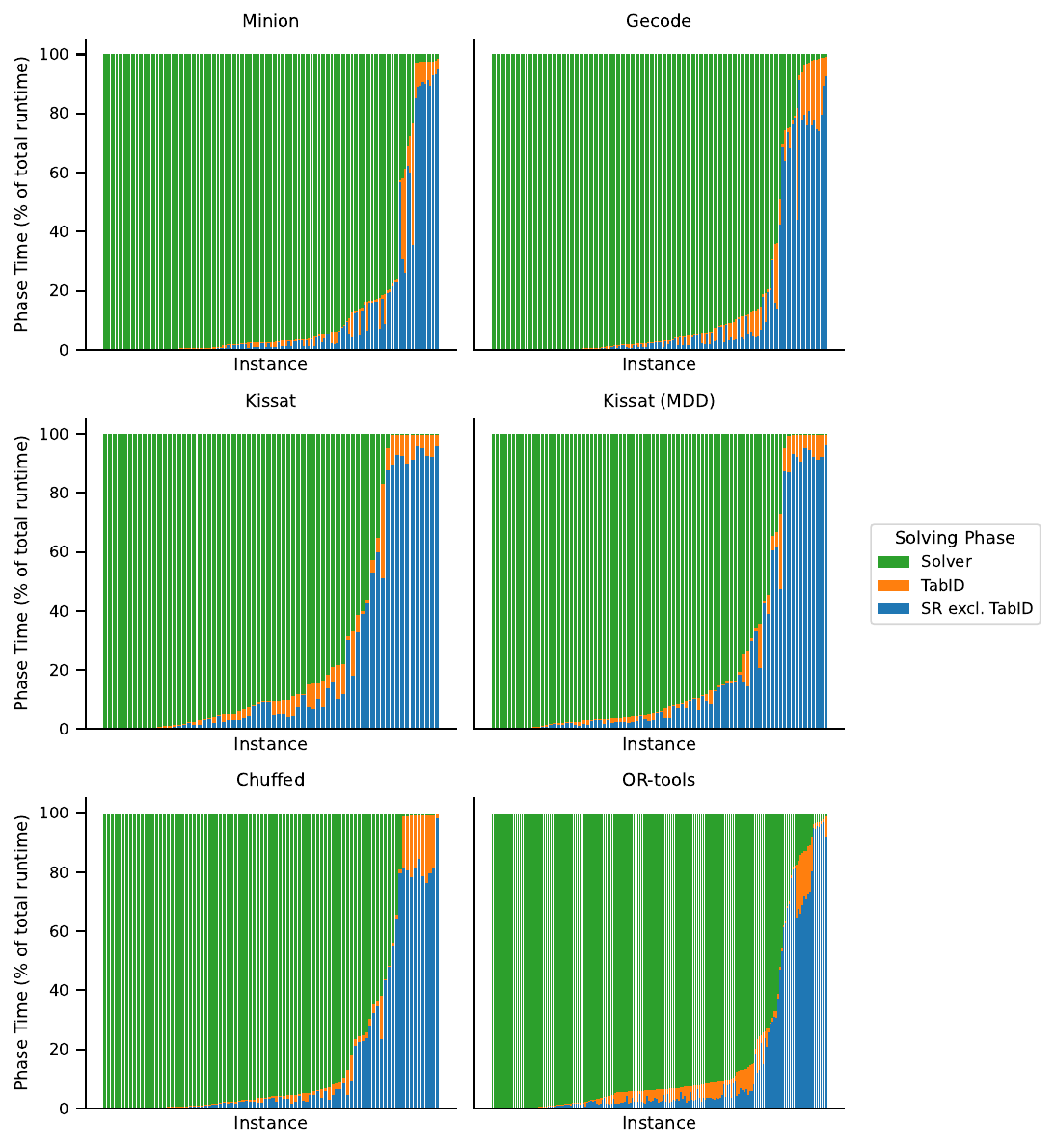}
\end{center}
\caption{\label{fig:phases}The time spent on the different phases of the solving process.  Instances are ordered by total \savilerow time along the horizontal axis; for each instance the runtime is split into three phases displayed vertically: firstly \savilerow's pre-processing excluding TabID, secondly the time spent on TabID, and finally the time taken by the target solver.  We show all instances from the featured problem classes in \Cref{sec:experimental-eval} whose total runtime is between 10 and 3600 seconds.}
\end{figure}


\section{Results with SAT Size Limit}\label{appendix:satsizelimit}

In \Cref{sub:satsizelimit} we describe the option to apply a \emph{SAT size limit} which avoids generating a table for a candidate expression if the resulting SAT encoding would be too large.
In this appendix we present the results of solving problems without TabID versus TabID with the SAT size limit applied; \Cref{fig:grid-dek-tabsatlim,fig:grid-results-a-satlim,fig:grid-results-b-satlim} show the comparison for the baseline problems as well as the new case studies (cross-reference with the results presented and analysed in \Cref{sec:exp-baseline,sec:exp-newproblems}).

In \Cref{fig:grid-dek-tabsatlim} we observe that the speedups with the SAT size limit applied is broadly similar for Minion, Gecode and OR-tools.  However, when we consider the results with Kissat, Kissat(MDD) and Chuffed, we see a marked improvement for the Block Party Metacube problem with almost all instances being solved faster than when TabID is turned off, and with the geometric mean of speedups rising from (0.52, 0.19, 1.29) to (2.10, 1.60, 7.60).  There are also improvements for the Maximum Density Still Life problem with the same three solvers.

\Cref{fig:grid-results-a-satlim} shows the results for the first batch of new case studies.  Here, the application of the SAT size limit does not change the results considerably in most cases.  The exception is the Killer Sudoku problem, where we see that with the SAT size limit there is a deterioration in performance with Minion and Gecode, but a big improvement with OR-tools.

Finally in \Cref{fig:grid-results-b-satlim} we have the results for the second batch of new case studies.  Once again the results here are very similar to the ones presented in \Cref{sec:exp-newproblems}.  The SAT size limit does have a slight negative impact on the Strong External Difference Families problem --- in this case some speedups were achieved by TabID even though the SAT encoding size of some of the tables are over the threshold imposed by the SAT size limit.

\section{Overhead of TabID}
\label{sec:tab-overhead}

We present a brief analysis of the time taken by TabID as compared to the other phases in the solving pipeline. The SAT size limit is switched off and \textit{nodeLimit} is set to 100,000 for this experiment. \Cref{fig:phases} shows how the total runtime is divided between \savilerow and the target solver.  The time spent in \savilerow is further split into two phases: TabID and everything else.  The middle TabID portion (shown in orange) is relatively small in the vast majority of cases -- sometimes it can take over half of the \savilerow time, but in those cases much longer is spent in the backend solver phase.

In most cases the majority of time is spent in the final solving phase carried out by the backend solver.  However, there are several instances where considerable time is spent in \savilerow to reformulate the problem ready for the target solver which then solves it almost instantly.

\bibliography{general,tabulation}
\bibliographystyle{theapaurl}

\end{document}